\documentclass{ieeeaccess}

\usepackage{cite}
\usepackage{amsmath,amssymb,amsfonts}
\usepackage{algorithm}
\usepackage{algorithmic}
\usepackage{graphicx}
\usepackage{textcomp}
\usepackage{color}
\usepackage{array}
\usepackage{bm}
\usepackage{mathtools}
\usepackage{tabularx}
\usepackage{arydshln}
\usepackage{multirow}

\DeclareMathOperator*{\argmin}{arg\,min}
\DeclareMathOperator*{\argmax}{arg\,max}
\DeclarePairedDelimiter\ceil{\lceil}{\rceil}

\def\BibTeX{{\rm B\kern-.05em{\sc i\kern-.025em b}\kern-.08em
    T\kern-.1667em\lower.7ex\hbox{E}\kern-.125emX}}

\begin{document}
\history{Date of publication xxxx 00, 0000, date of current version May 19, 2021.}
\doi{10.1109/ACCESS.2017.DOI}

\title{Quantifying the Complexity of Standard Benchmarking Datasets for Long-Term Human Trajectory Prediction}
\author{\uppercase{Ronny Hug\authorrefmark{1}, Stefan Becker\authorrefmark{1}, Wolfgang H\"ubner\authorrefmark{1}, and Michael Arens\authorrefmark{1}}}
\address[1]{Fraunhofer IOSB, Ettlingen, Fraunhofer Institute of Optronics, System Technologies and Image Exploitation, 76275 Ettlingen, Germany (e-mail: firstname.lastname@iosb.fraunhofer.de)}
\tfootnote{Fraunhofer IOSB is a member of the Fraunhofer Center for Machine Learning.}

\markboth
{Hug \headeretal: Quantifying the Complexity of Standard Benchmarking Datasets for Long-Term Human Trajectory Prediction}
{Hug \headeretal: Quantifying the Complexity of Standard Benchmarking Datasets for Long-Term Human Trajectory Prediction}

\corresp{Corresponding author: Ronny Hug (e-mail: ronny.hug@iosb.fraunhofer.de).}

\begin{abstract}

Methods to quantify the complexity of trajectory datasets are still a missing piece in benchmarking human trajectory prediction models.
In order to gain a better understanding of the complexity of trajectory prediction tasks and following the intuition, that more complex datasets contain more information, an approach for quantifying the amount of information contained in a dataset from a prototype-based dataset representation is proposed.
The dataset representation is obtained by first employing a non-trivial spatial sequence alignment, which enables a subsequent learning vector quantization (LVQ) stage.
A large-scale complexity analysis is conducted on several human trajectory prediction benchmarking datasets, followed by a brief discussion on indications for human trajectory prediction and benchmarking.

\end{abstract}


\titlepgskip=-15pt

\maketitle

\section{Introduction} 
With the emergence of autonomous vehicles and advances in the field of intelligent robots in general, the task of human trajectory prediction gained a significant amount of research interest in recent years.
Besides more classical, physics-based prediction approaches, e.g. building on the Kalman filter \cite{kalman1960} or the social forces model \cite{helbing1995social}, a range of deep learning approaches have been proposed to tackle the problem.
The most common deep learning models either build around long short-term memory networks (e.g. \cite{becker2018red}), convolutional neural networks (e.g. \cite{Nikhil_ECCVW_2018}), generative adversarial networks (e.g. \cite{gupta2018social}) or transformers (e.g. \cite{giuliari2020transformer}) and vary in contextual cues considered for prediction.
Commonly used contextual cues include social (e.g. \cite{alahi2016social}\cite{huang2019stgat}) and environmental (e.g. \cite{sadeghian2019sophie}\cite{bisagno2021embedding}) cues.
For a comprehensive overview of existing human trajectory prediction approaches, the reader may be referred to \cite{rudenko2019human}\cite{rasouli2020deep}\cite{kothari2020human}.

Inherent to prediction model development is the need for proper benchmarks aimed at measuring a model's prediction performance.
Due to the direct relation between dataset complexity and model capacity, creating a not too simple or too hard-to-solve benchmark for human trajectory prediction is still a difficult task.
On large datasets, even simple models overfit, while in other cases the prediction performance is poor on individual samples, even for high capacity models.
Here, one of the difficulties is the open question of how the complexity of a given dataset for trajectory prediction can be quantified.
As a consequence, current attempts in standardized benchmarking originate from heuristics or experience-based criteria when assembling the data basis.
Recent examples are the \emph{TrajNet} challenge \cite{sadeghiankosaraju2018trajnet} or its extension \emph{TrajNet++} \cite{rudenko2019human}\cite{kothari2020human}.

Currently, in human trajectory prediction, the analysis of benchmarking data only takes on a minor role and focuses on a specific aspect of the data.
The most common subject for analysis is the existence of social interaction resulting in non-linear behavior with respect to motion.
For such analyses, social force models and collision avoidance methods \cite{robicquet2016learning} or deviations from regression fits \cite{becker2018red} \cite{zhang2020srlstm} are employed for example.
Such methods are also used in TrajNet++ for splitting up the benchmark into interaction and non-interaction tasks.
Besides that, basic analyses include dissecting velocity profiles or positional distributions \cite{hug2017reliability}.  
Following that, there is still a lack of approaches trying to analyze datasets as a whole, with the goal of quantifying the overall complexity of the dataset.

When targeting a qualitative analysis of data complexity, a common approach are low-dimensional embeddings for data visualization, like for example \emph{t-SNE} \cite{maaten2008tsne} or variations of \emph{PCA} \cite{jolliffe2016pca}.
While such approaches are viable for non-sequential, high-dimensional data, a prototype-based clustering approach seems more viable for sequential data.
This is especially true for trajectory datasets, where each dataset can be reduced to a small number of prototypical sub-sequences specifying distinct motion patterns, where each sample can be assumed to be a variation of these prototypes.
Additionally, in the context of statistical learning, the complexity of a dataset can be closely related to the entropy of a given dataset.
Measuring the dataset entropy, i.e. the amount of information contained in a dataset, is still an open question in the context of trajectory datasets gaining interest recently \cite{hug2020shortnote}\cite{amirian2020opentraj}.

Towards this end, an approach for estimating an entropy-inspired measure, the pseudo-entropy, building on a dataset decomposition generated by an adequate pre-processing and clustering method is proposed. 
For decomposing a given dataset into clusters of distinct velocity-agnostic motion patterns, a spatial alignment\footnote{Not to be confused with a temporal sequence alignment} step followed by vector quantization is applied. 
Given the dataset decomposition, the pseudo-entropy is estimated by analyzing the prediction performance of a simple trajectory prediction model when gradually enriching its training data with additional motion patterns.
This paper is an extension of \cite{hug2020shortnote} and focuses mainly on dataset complexity.
The main contributions are:
\begin{enumerate}
	\item A learning and heuristics-based approach for finding a velocity-agnostic, prototype-based representation of trajectory datasets.  
	\item An approach for estimating trajectory dataset complexity in terms of an entropy-inspired measure.
	\item A coarse complexity-based ranking of standard benchmarking datasets for human trajectory prediction.
\end{enumerate}

The paper is structured as follows.
Sections \ref{sec:da} and \ref{sec:vq} present a data pre-processing approach necessary for the actual dataset entropy estimation detailed in section \ref{sec:entropy}.
In addition to a coarse dataset ranking, the evaluation section \ref{sec:evaluation} discusses the ranking, as well as the approach and methods used throughout this paper.
Further, some interesting findings resulting from the analysis are discussed.
Section \ref{sec:summary} summarizes the paper, lists potential implications for the state of benchmarking and gives a brief discussion on potential future research directions.

For convenience, several definitions and notations used throughout the paper are listed below:
\begin{itemize}
	\item A trajectory $X$ is defined as an ordered set\footnote{strictly speaking it is not a \emph{true} mathematical set, as it might contain duplicates.} of points $\{\mathbf{x}_1, ..., \mathbf{x}_N\}$.
	\item The \emph{length} of a (sub-)trajectory $X$ always refers to its cardinality $|X|$, rather than the spatial distance covered.
	\item The \emph{distance} between two trajectories $X$ and $Y$ of the same length $|X| = |Y|$ is defined as $d_{tr}(X, Y) = \sum_{j=1}^{|X|} \| \mathbf{x}_j - \mathbf{y}_j \|_2$. 
	\item The number of samples, the trajectory length and the number of prototypes are denoted as $N$, $M$ and $K$, respectively. For indexing, $i$, $j$ and $k$ are used.
	\item The $q$-quantile, with $q \in [0, 1]$, of a set of numbers $\{\cdot\}$ is denoted as $Q_q(\{ \cdot \})$.
\end{itemize}

\section{Spatial Sequence Alignment}
\label{sec:da}
With the goal of reaching a velocity-agnostic, prototype-based trajectory dataset representation in mind, the trajectory alignment approach proposed in this section fulfills two integral roles as a pre-processing for the subsequent clustering step. 
On the one hand, it aligns the data, such that similar patterns are pooled together.
On the other hand, it removes variations in velocity among trajectories, therefore generating a dataset with normalized velocity.
This, in turn, is essential in obtaining a velocity-agnostic dataset representation.
In addition, velocity normalization ensures that similar motion patterns, which only vary in velocity, i.e. in scale, can be pooled together.

Given a set of trajectories (\emph{samples}) $\mathcal{X} = \{X_1, ..., X_N\}$, as sequences of $M$ subsequent points $X_i = \{\mathbf{x}^i_1, ..., \mathbf{x}^i_M\}$, each sample is first normalized by moving it into an arbitrary reference frame and scaling it to unit length:
\begin{align}
	X^{norm}_i = \left\{\frac{\mathbf{x}^i_j - \bar{\mathbf{x}}}{\mathbf{x}^i_M - \mathbf{x}^i_1} \mid j \in [1,M] \right\}.
\end{align}
Here, $\bar{\mathbf{x}}$ is the centroid of a trajectory.
It has to be noted that this normalization solely serves the purpose of moving all samples into a common value range and therefore it is not a good normalization in terms of pooling similar samples.
Then, all samples are aligned relative to a single learned prototype $\hat{Y} = \left\{\mathbf{\hat{y}}_1, \mathbf{\hat{y}}_2, ..., \mathbf{\hat{y}}_M \right\}$ by using similarity transformations, which are retrieved from a regression model $\phi: X \rightarrow \{\mathbf{t},\alpha,s\}$ with translation $\mathbf{t}$, rotation angle $\alpha$ and scale $s$.
$\hat{Y}$ and $\phi$ are learned by minimizing the mean squared error between each aligned sample $X^\phi_i = \phi(X^{norm}_i)$ and the prototype $\hat{Y}$
\begin{align}
	\label{eq:loss_align}
	\mathcal{L}_{\mathrm{align}}(\phi(X^{norm}_i), \hat{Y}) = \frac{1}{M} \sum^M_{j=1} \|\mathbf{x}^i_j - \mathbf{\hat{y}}_j\|^2_2 
\end{align}
using stochastic gradient descent.
This is different from linear factor models, where the whole training set has to be considered.
With respect to equation \ref{eq:loss_align} and the similarity transformation, the trivial solution that maps all samples onto the zero-vector has to be avoided.
A brute force approach to this problem is to enforce a minimum scale for the prototype. 
These steps result in a minimum variance alignment of all samples with respect to the learned prototype.
Further, by learning the prototype and the transformation concurrently, the prototype adapts to the most dominant motion pattern and the normalized data is aligned accordingly. 

\begin{figure}[htb]
	\centering
	\includegraphics[width=0.95\columnwidth]{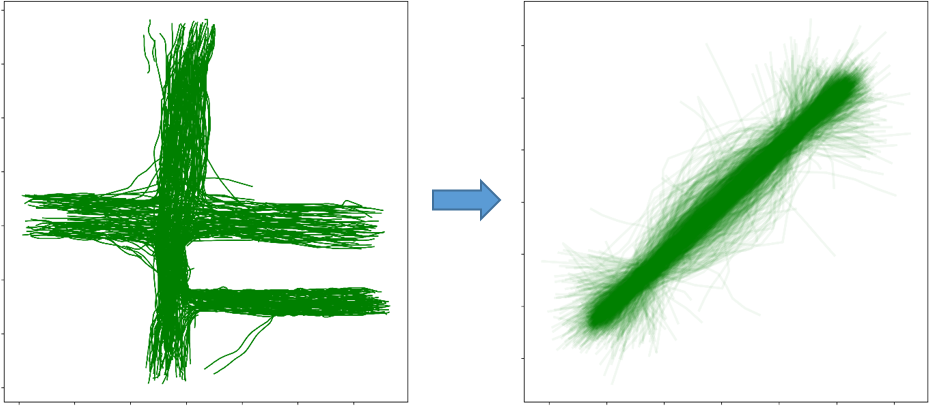}
	\caption{Example for a resulting alignment (right) of a given dataset (left; \emph{hyang} scene taken from the Stanford Drone Dataset \cite{robicquet2016learning}).}
	\label{fig:align_ex}
\end{figure}

An exemplary result of this alignment approach is depicted in figure \ref{fig:align_ex}.
By aligning all samples with a single prototype, aligned samples have a common orientation and form clusters of similar samples.

\section{Learning Vector Quantization}
\label{sec:vq}
Clustering approaches can be applied after the spatial alignment, as it distributes random errors homogeneously over the sequence and exposes clusters of motion patterns.
In the landscape of clustering approaches, there exists a wide range of approaches to choose from, ranging from simple established approaches (e.g. \emph{k-means}\cite{lloyd1982least}, \emph{DBSCAN}\cite{ester1996density} or \emph{learning vector quantization} \cite{kohonen2001lvq}) to more sophisticated or specialized neural models (e.g. \cite{xie2016unsupervised}\cite{caron2018deep}\cite{peng2019deep}).
In the context of this paper, the choice of the clustering approach itself only plays a minor role, thus a learning vector quantization (LVQ) approach is employed, as it can directly be integrated into a deep learning framework.
For a more comprehensive review on clustering approaches, the reader may be referred to recent surveys, for example \cite{xu2015comprehensive} or \cite{min2018survey}. 

The resulting pipeline corresponds to the encoder part of an auto-encoding architecture for representation learning inspired by \cite{oord2017neural}, which is capable of learning meaningful representations.
Here, aligned samples are mapped onto $K$ prototypes\footnote{These are distinct from the alignment prototype $\hat{Y}$} $\mathcal{Z} = \{Z_1, ..., Z_K\}$, with $Z_k = \{\mathbf{z}^k_1, ..., \mathbf{z}^k_M\}$, in quantized space.
This results in a concise set of prototypes representing the given dataset.
The prototypes are learned by minimizing the mean squared error between the aligned samples $\mathcal{X}^{\phi} = \{X^\phi_1, ..., X^\phi_N\}$ and the respective closest prototype $Z_{z(i)}$ in quantized space:
\begin{align} 
	\mathcal{L} = \underbrace{\frac{1}{N} \sum^N_{i=1} d_{tr}(X^\phi_i, Z_{z(i)})}_{\mathcal{L}_{LVQ}} + \gamma L_{reg},
\end{align}
where $L_{reg}$ is the regularization term, which is discussed in section \ref{ss:regularization}.  
The index of the closest prototype for a sample $X^\phi_i$ is determined by
\begin{align}
	z(i) = \argmin_k d_{tr}(X^\phi_i, Z_k).
\end{align}
Note that due to the fixed trajectory length, the mean squared error is a suitable similarity measure.
If the length would vary, a more sophisticated measure would be necessary \cite{quehl2017good}.

Using $\mathcal{L}$ for learning the LVQ parameters, two aspects have to be considered:
\begin{enumerate}
	\item As the value for $K$ is unknown a priori, it should in general be chosen larger than expected.
	\item Due to the winner takes all strategy, $\mathcal{L}_{LVQ}$ only updates prototypes that have supporting samples.
\end{enumerate}
In order to achieve consistent training and quantization results under these conditions, the following sections present approaches for initialization (\ref{ss:initialization}), regularization (\ref{ss:regularization}) and refinement (\ref{ss:refinement}).
While the initialization and regularization primarily focus on 2), the refinement, build upon the expected results using the proposed initialization and regularization approaches, focuses on 1).

For describing these approaches, the \emph{support of a prototype} plays an integral role. 
The support $\pi(\mathcal{Z})_k$ of the aligned dataset $\mathcal{X}^\phi$ for each prototype $Z_k \in \mathcal{Z}$ is aggregated in  
\begin{align}
	\label{eq:support}
	\begin{split}
		\pi(\mathcal{Z}) &= \left\{ \sum^{N}_{i = 1} \mathbf{1}_k(i) \mid k \in [1, \lvert \mathcal{Z} \rvert] \right\}, \\ 
		\mathrm{with} & \\
		\mathbf{1}_k(i) &= \begin{cases} 1 & \mathrm{if}~z(i) = k \\ 0 & \mathrm{else} \end{cases}.
	\end{split}
\end{align}

\begin{figure}[htb]
	\centering
	\includegraphics[width=0.95\columnwidth]{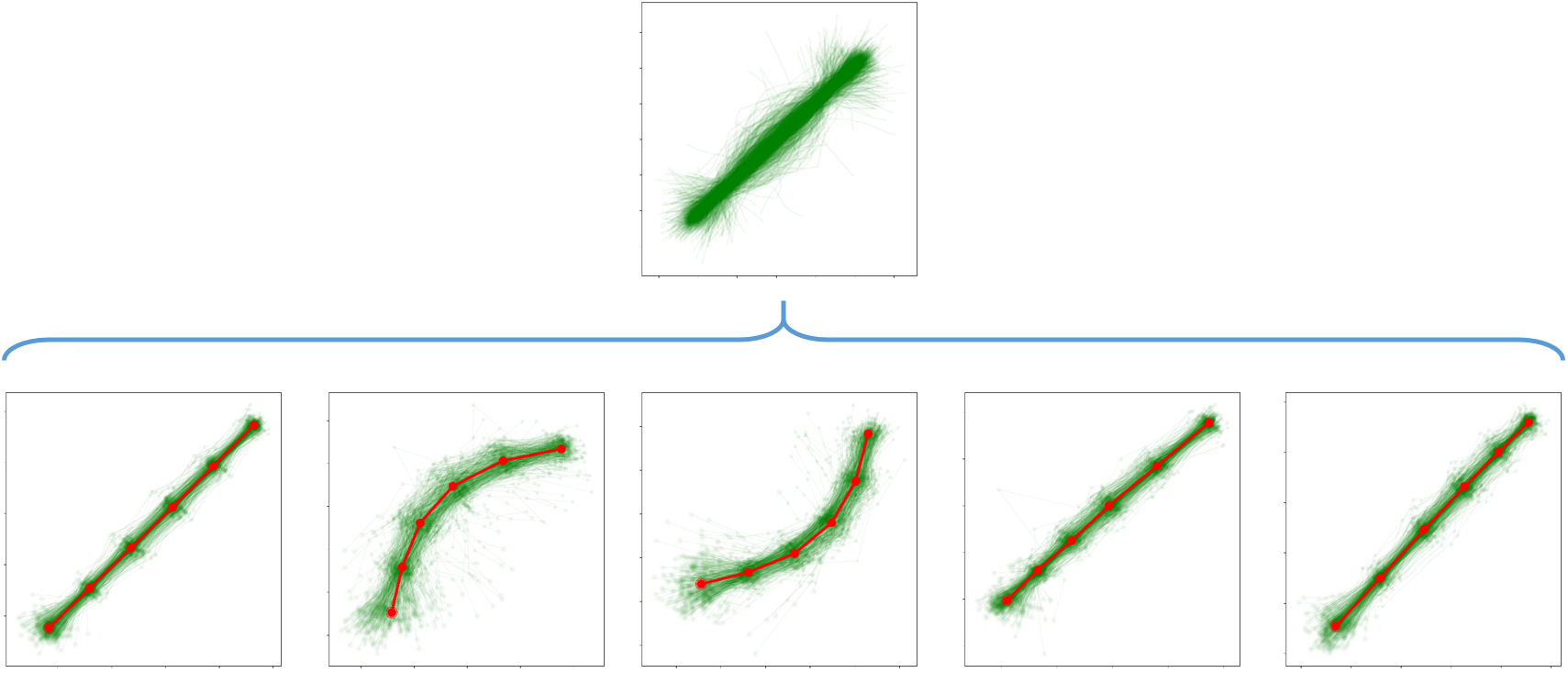}
	\caption{Possible set of prototypes (red) for a given aligned dataset (top row). The prototypes represent different motion patterns (from left to right): Constant velocity, curvilinear motion, acceleration and deceleration.}
	\label{fig:protos_ex}
\end{figure}

A resulting set of prototypes using the approach described in this section and following subsections, for the dataset shown in figure \ref{fig:align_ex}, is depicted in figure \ref{fig:protos_ex}.
It can be seen, that the prototypes cover a certain range of motion patterns: constant velocity, curvilinear motion, acceleration and deceleration.

\subsection{Initialization}
\label{ss:initialization}
The main objective of the initialization step is two-fold.
On the one hand, the number of out-of-distribution prototypes should be reduced in the initial set of prototypes $\mathcal{Z}_{\mathrm{init}}$. 
On the other hand, $\mathcal{Z}_{\mathrm{init}}$ has to be spread across the data $X^\phi$, in order to identify different motion patterns more consistently.

\begin{figure}[htb]
	\centering
	\includegraphics[width=0.95\columnwidth]{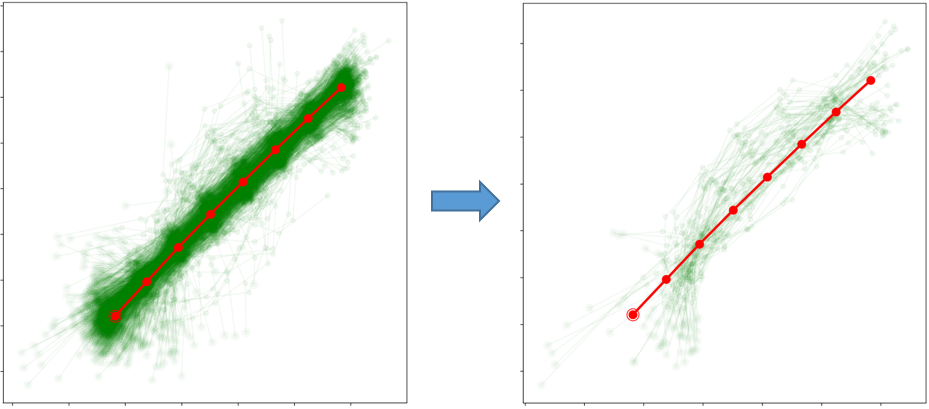}
	\caption{Left: Aligned dataset $\mathcal{X}^\phi$ (green) and alignment prototype $\hat{Y}$ (red). 
		Right: Subset $\mathcal{X}'$ for $q_{\mathrm{ilow}} = 0.9$ and $q_{\mathrm{ihigh}} = 0.95$.}
	\label{fig:init_ex}
\end{figure}

Taking this into account, the alignment prototype $\hat{Y}$ is set as the first prototype $Z_1$, as it should resemble the most dominant motion pattern.
Under the assumption that other relevant motion patterns are dissimilar to $\hat{Y}$, a Forgy initialization \cite{pena1999empirical} is applied for initializing the remaining $K-1$ prototypes.
Accordingly, the remaining prototypes are randomly selected from a subset $\mathcal{X}' \subset \mathcal{X}^\phi$, where $\mathcal{X}'$ contains all samples $\mathcal{X}^\phi_i$ with $\tau_{\mathrm{ilow}} < d_{tr}(X^\phi_i, \hat{Y}) < \tau_{\mathrm{ihigh}}$.
The thresholds $\tau_{\mathrm{ix}} = Q_{q_{\mathrm{ix}}}(\{ d_{tr}(X^\phi_i, \hat{Y}) \mid X^\phi_i \in \mathcal{X}^\phi \})$ are defined as the $q_{\mathrm{ix}}$-quantiles of all sample distances with respect to $\hat{Y}$.
An upper bound $\tau_{\mathrm{ihigh}}$ is employed to reduce the risk of initializing a prototype with out-of-distribution samples from $\mathcal{X}^\phi$.
Depending on the choices for $q_{\mathrm{ilow}}$ and $q_{\mathrm{ihigh}}$, there might not be enough samples to choose from ($\lvert \mathcal{X}' \rvert < K-1$).
In this case, $q_{\mathrm{ilow}}$ can be gradually reduced until $\lvert \mathcal{X}' \rvert \geq K-1$.
An example for $\mathcal{X}$ and $\mathcal{X}'$ with $q_{\mathrm{ilow}} = 0.9$ and $q_{\mathrm{ihigh}} = 0.95$ is given in figure \ref{fig:init_ex}.

\subsection{Regularization}
\label{ss:regularization}
While the initialization helps in increasing the average support for each prototype\footnote{Compared to a simple random initialization}, some out-of-distribution samples\footnote{Outliers or trajectories with annotation errors} might be assigned to individual prototypes, resulting in little support from other samples.

To ensure optimization of all prototypes, a regularization term $L_{reg}$ is employed, which is set to move out-of-distribution prototypes closer to more relevant samples or clusters of samples.
Following this, different definitions for $L_{reg}$ can be used.
On the one hand, $L_{reg}$ could move all prototypes $Z_k \in \mathcal{Z} \setminus Z_*$ towards the most supported prototype $Z_* = Z_{k_*}$, where $k_* = \argmax_k \pi(\mathcal{Z})_k$:
\begin{align}
	L_{reg} = \frac{1}{K - 1} \sum_{Z_k \in \mathcal{Z} \setminus Z_*} d_{tr}(Z_k, Z_*).
\end{align}
Under ideal conditions, $Z_*$ should be equal to the alignment prototype $\hat{Y}$, which roughly represents the overall mean of the dataset, and $L_{reg}$ behaves accordingly.
In practice, however, this assumption might not hold due to noise, increasing unpredictability of the optimization.
Hence, in the following $L_{reg}$ is defined to move all prototypes towards the global mean by minimizing the global error
\begin{align}
	L_{reg} = \frac{1}{N \cdot K} \sum_{i=1}^{N}\sum_{k=1}^{K} d_{tr}(X_i, Z_k).
\end{align}
Intuitively, by choosing an appropriate value for the regularization weight $\gamma$, this definition of $L_{reg}$ moves low-support prototypes in more reasonable areas within quantized space and the winner takes all loss function $\mathcal{L}_{LVQ}$ keeps them within range of relevant sample clusters.
Additionally, when $K$ is too large, superfluous prototypes are very similar after optimization.

As a side-note, very imbalanced prototype sets, in terms of many low-support prototypes, can also be measured by the perplexity score
\begin{align}
	\begin{split}
		&\mathcal{P}_\mathcal{Z} = \exp\left\{ -\sum_{k=1}^K \pi_{\mathrm{norm}}(\mathcal{Z})_k \odot \log\{ \pi_{\mathrm{norm}}(\mathcal{Z})_k \} \right\}, \\
		&\mathrm{with} \\
		&\pi_{\mathrm{norm}}(\mathcal{Z})_k = \frac{\pi(\mathcal{Z})_k}{\sum_{k=1}^{K} \pi(\mathcal{Z})_k}. 
	\end{split}
	\label{eq:perplexity}
\end{align}
Due to $\mathcal{P}_\mathcal{Z}$ not being directly derived from $\mathcal{Z}$, it is not a good term for optimization, and thus for $L_{reg}$. 
Nevertheless, $\mathcal{P}_\mathcal{Z}$ can be used later on when assessing the complexity of a dataset.

\subsection{Refinement}  
\label{ss:refinement}
Finally, a heuristic refinement scheme, building on the expected results when using the regularization approach presented in section \ref{ss:refinement}, is employed in order to remove unnecessary prototypes when $K$ was too large.
The refinement step consists of two phases.
In the first phase, low-support prototypes are removed from $\mathcal{Z}$ by using a dataset-dependent threshold $\tau_{\mathrm{phase1}} = \ceil{\epsilon_{\mathrm{phase1}} \cdot |X|}$:
\begin{align} 
	\mathcal{Z}' = \mathcal{Z} \setminus \left\{ Z_k \mid \pi(\mathcal{Z})_k < \tau_{\mathrm{phase1}} \right\}. 
\end{align}

\begin{figure}[htb] 
	\centering
	\includegraphics[width=0.95\columnwidth]{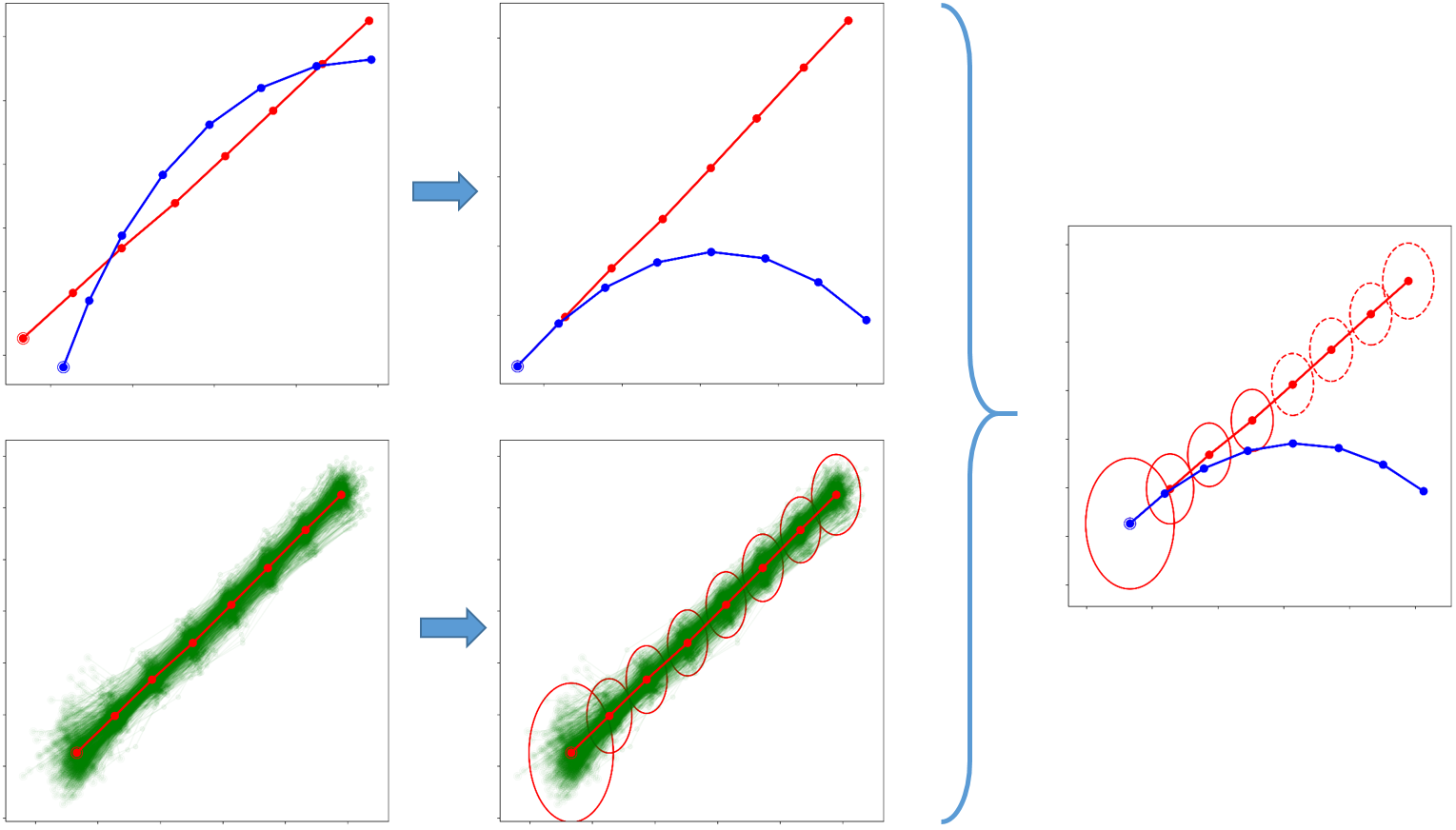}
	\caption{Example of assessing prototype similarity in the second phase of the refinement scheme.
		The first row shows the alignment of a prototype $Z_k$ (blue) with the highest-support prototype $Z_*$ (red),
		The second row shows how the maximum distance per-point is estimated for determining similarity.
		In this example, as seen on the right, $Z_k$ is only within similarity range for $4$ of $8$ points, thus it is determined as dissimilar when choosing an overlap factor $\epsilon_{\mathrm{phase2}} > 0.5$.} 
	\label{fig:refinement_ex}
\end{figure}

The second phase revolves around removing prototypes similar to the most supported prototype $Z_*$.  
It is assumed, that $Z_*$ is close to the global mean of the dataset. 
This implies, that superfluous prototypes are driven towards $Z_*$ because of the global mean regularization, allowing to detect and remove these prototypes.
For assessing similarity, prototypes $Z_k \in \mathcal{Z} \setminus \{Z_*\}$ are first aligned with $Z_*$ in terms of their starting points $\mathbf{z}^k_1 = \mathbf{z}^*_1$ and initial orientations $\mathbf{z}^k_2 - \mathbf{z}^k_1 = \mathbf{z}^*_2 - \mathbf{z}^*_1$.
An aligned prototype $Z'_k$ is then considered as similar to $Z_*$ when at least $\epsilon_{\mathrm{phase2}} \cdot 100~\%$ of its points are in close proximity to respective points of $Z_*$: 
\begin{align}
	\begin{split}
		\mathrm{sim}(Z'_k, Z_*) &= \begin{cases} 1 & \mathrm{if}~|\mathcal{S}(Z'_k, Z_*)| \geq \epsilon_{\mathrm{phase2}} \cdot \lvert Z_* \rvert \\ 0 & \mathrm{else} \end{cases}, \\ 
		\mathrm{with} & \\
		\mathcal{S}(Z'_k, Z_*) &= \left\{ \mathbf{z}^k_j \mid \| \mathbf{z}^k_j - \mathbf{z}^*_j \|_2 < \tau(Z_*)_j, j \in [1, M] \right\} \\
		\tau(Z_*)_j &= Q_{0.99}\left( \left\{ \| \mathbf{z}^*_j - \mathbf{x}_j \|_2 \mid \mathbf{x}^i_j \in X_i,~X_i \in \mathcal{X}^\phi \right\} \right).
	\end{split}
\end{align}
$\tau(Z_*)_j$ is the per-point distance threshold of the $j$'th trajectory point calculated from the supporting samples $\{X_i \in \mathcal{X}^\phi \mid z(i) = k_* \}$ of $Z_*$.
The $0.99$-quantile is used instead of the maximum to exclude outliers in the data.
A visual example for determining similarity is given in figure \ref{fig:refinement_ex}.

\section{Estimating Trajectory Dataset Entropy}
\label{sec:entropy}
This section discusses an attempt in moving towards enabling a thorough complexity analysis of human trajectory prediction benchmarking datasets.
While previous work (e.g. \cite{robicquet2016learning}, \cite{hug2017reliability}, \cite{becker2018red}) focuses on statistics directly derived from the datasets, like histograms or deviations from linear prediction, the approach proposed in the following relies on a dataset decomposition $\mathcal{X}^\phi_{decomp} = \{\mathcal{X}_1, ..., \mathcal{X}_k\}$ learned by the LVQ model introduced in section \ref{sec:vq}, i.e. the clusters $\mathcal{X}_k \in \mathcal{X}^\phi$ assigned to each prototype $Z_k \in \mathcal{Z}$ \emph{after} refinement.
This decomposition is used for an effort in estimating an entropy-inspired measure for a given trajectory dataset $\mathcal{X}$, declared as the \emph{pseudo-entropy} $\text{H}_{pseudo}(\mathcal{X})$.
Given a dataset, it is assumed, that a high information content yields a high entropy value, which finally gives a direct quantification of complexity.
Compared to the \emph{OpenTraj} approach described in \cite{amirian2020opentraj}, which was developed independently at the same time and mainly focuses on analyzing raw trajectory data, the approach presented in this paper focuses on the analysis of more abstract trajectory data in alignment space and clusters of motion patterns extracted from aligned data. 

As an initial proof of concept, $\text{H}_{pseudo}(\mathcal{X})$ is estimated using the change in prediction performance of a simple machine learning-based trajectory prediction model when gradually increasing the amount of information in the training data.
The amount of information in the training data can be controlled to a certain extent by using the dataset decomposition in alignment space generated by the LVQ model.
In general, most current trajectory datasets are strongly biased towards linear motion patterns, thus other, more complex patterns are less relevant and carry more information.
Following this, by ordering the clusters in $\mathcal{X}^\phi_{decomp}$ by decreasing relevance, a training set initially consisting only of $\mathcal{X}_1 \in \mathcal{X}^\phi_{decomp}$ can be gradually enriched with information by first adding $\mathcal{X}_2$, then $\mathcal{X}_3$ and so on.
Then, a simple prediction model $\mathcal{M}$ is trained on each composed training dataset. 
In this paper, $\mathcal{M}$ is comprised of a learned linear input transformation
\begin{align}
	\mathcal{M}(X_f) = WX_f + b,
\end{align}
where $W$ and $b$ are learned from data and $X_f$ is a flattened vector given by concatenating all points of a trajectory $X$.
This model is assumed to be able to model each relevant motion pattern in isolation.
Due to the lack of modeling capacity, it is expected that the prediction error increases with the increase of information present in the data.
Finally, $\text{H}_{pseudo}(\mathcal{X}^\phi)$ is calculated by counting significant prediction error increases as depicted in algorithm \ref{algo:estimate_entropy}.
\begin{algorithm}
	\begin{algorithmic}
		\REQUIRE Decomposed, sorted datasets $\mathcal{X}^\phi_{decomp}$
		\STATE $\text{PseudoEntropy} \leftarrow 0$
		\FOR{$k \in \{1, ..., K\}$}
		\STATE $\mathcal{X}_{train} = \bigcup_{i=1}^{k} \mathcal{X}_i,~\mathcal{X}_i \in \mathcal{X}^\phi_{decomp}$
		\STATE $\mathcal{M}_k \leftarrow \text{estimateParameters}(\mathcal{X}_{train})$ 
		\IF{$k > 1$} 
		\STATE $\text{errs}_{k-1} \leftarrow \text{calcPredictionErrorsPerSample}(\mathcal{M}_{k-1})$
		\STATE $\text{errs}_{k} \leftarrow \text{calcPredictionErrorsPerSample}(\mathcal{M}_{k})$
		\IF{$\text{SignificantDifference}(\text{errs}_{k-1}, \text{errs}_{k}) \land \text{mean}(\text{errs}_{k}) > \text{mean}(\text{errs}_{k-1})$}
		\STATE PseudoEntropy++
		\ENDIF
		\ENDIF
		\ENDFOR
		\RETURN PseudoEntropy
	\end{algorithmic}
	\caption{Estimate Dataset Pseudo-Entropy}
	\label{algo:estimate_entropy}
\end{algorithm}

\section{Evaluation} 
\label{sec:evaluation}
This section starts with a setup common to all experiments, followed by a qualitative evaluation of the simple prediction model described in section \ref{sec:entropy} and the LVQ model for dataset decomposition.
Next, a coarse ranking of standard benchmarking datasets for long-term human trajectory prediction based on pseudo-entropy is given.
The section closes with a discussion on the approach and methodology itself, other possible factors contributing to dataset complexity and interesting findings.

Evaluations are conducted on scenes taken from the following frequently used benchmarking datasets: BIWI Walking Pedestrians (\cite{pellegrini2009biwi}, abbrev.: \emph{biwi}), Crowds by example (also known as the UCY dataset, \cite{lerner2007crowds}, abbrev.: \emph{crowds}) and the Stanford Drone Dataset (\cite{robicquet2016learning}, abbrev.: \emph{sdd}).
Besides being typically used for evaluating human trajectory prediction models in the literature, the original TrajNet challenge was built around these datasets.
The scenes in the datasets are denoted as \emph{Dataset: Scene Recording}, e.g. recording $01$ of the \emph{zara} scene in the \emph{crowds} dataset is denoted as \emph{crowds:zara01}.
Note that for \emph{sdd}, different recordings of the same scene do not necessarily capture the same campus area (but there might be some overlap).
An overview of statistical details of the datasets is given in appendix \ref{app:datasets}.  

For trajectory prediction tasks targeted in this section, trajectories are split in half in order to obtain observation and target sequences.
The prediction error is reported in terms of the average displacement error.

\subsection{Setup}
First, the datasets are augmented to have a common sample frequency.
The \emph{biwi} and \emph{crowds} scenes already have the same sample frequency of $2.5$ samples per second, thus the sample frequency of all the \emph{sdd} scenes is adjusted accordingly.

Next, as the prototype-based representation only works with trajectories of the same length, an appropriate sequence length has to be chosen for each dataset in the evaluation.
The most commonly used sequence length in recent benchmarks is $M=20$ ($8$ for observation and $12$ for predicting) points per trajectory.
Setting $M=20$ for all datasets might, however, lead to smaller datasets having only few trajectories left for learning the LVQ, as the average trajectory length varies greatly between datasets.
Because of this, the $q$-quantile of trajectory lengths per dataset is chosen, i.e. a common but dataset-dependent sequence length.
After choosing $M$, the training datasets are assembled by collecting all possible (sub-)trajectories with length $M$ from each respective dataset, in order to provide as much data as possible.
Additionally, for achieving a more meaningful result, the average of multiple sequence lengths, i.e. time scales, is used for calculating the pseudo-entropies.
Thus, $q = 0.1$ and $q = 0.25$ are used for evaluation, ensuring that a greater portion of the dataset remains while removing less interesting trajectories in terms of long-term trajectory prediction.

Then, trajectories not exceeding a dataset-dependent minimum speed\footnote{The average distance between subsequent trajectory points.} $s_{\mathrm{min}}$ are filtered.
The reason for this is twofold.
First, statistical models are worse in modeling trajectories of slow-moving persons, as their behavior becomes less predictable \cite{hasan2019forecasting}.
Second, and as a consequence, these trajectories generally do not contain viable motion patterns to extract.
For this evaluation, the minimum speed is calculated heuristically for a given training dataset $\mathcal{X}$ containing all possible (sub-)trajectories of length $M$:  
\begin{align}
	\begin{split}
		s_{\mathrm{min}} &= \frac{\max_i \mathrm{m}_{\mathrm{speed}}(i) - \min_i \mathrm{m}_{\mathrm{speed}}(i)}{M} \\
		\mathrm{with} & \\
		\mathrm{m}_{\mathrm{speed}}(i) &= \frac{1}{M - 1} \sum_{j=2}^{M} \| \mathbf{x}^i_j - \mathbf{x}^i_{j - 1} \|_2
	\end{split}
\end{align}
Here, $\mathrm{m}_{\mathrm{speed}}(i)$ denotes the average speed of the $i$'th trajectory $X_i \in \mathcal{X}$.

Finally, for each dataset and sequence length, the training of the alignment and LVQ networks are run $10$ times.
The number of initial prototypes is set to $K = 10$ for all datasets.
If not stated otherwise, the refinement parameters are set to $\epsilon_{\mathrm{phase1}} = 0.04$ and $\epsilon_{\mathrm{phase2}} = 0.9$.

\subsection{Simple Prediction Model Capabilities} 
\begin{figure}[h]
	\setlength{\tabcolsep}{3pt}
	\centering
	\begin{tabular}{ccc}
		\includegraphics[width=0.31\columnwidth]{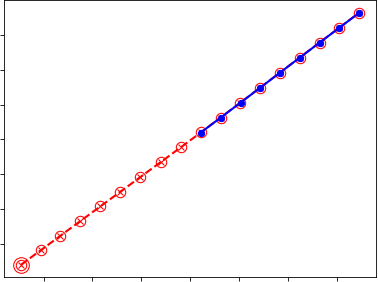} &
		\includegraphics[width=0.31\columnwidth]{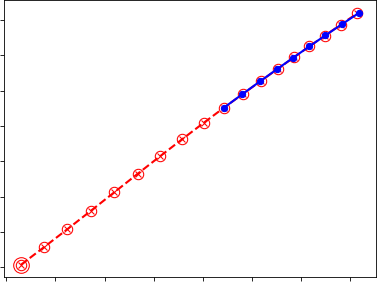} &
		\includegraphics[width=0.31\columnwidth]{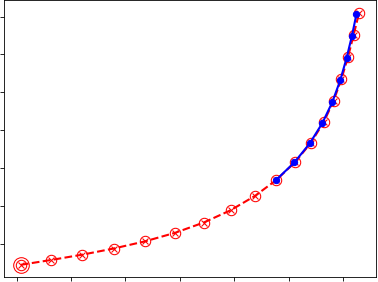} \\
		a. & b. & c. 
	\end{tabular}
	\caption{Prediction results of a learned linear transformation (blue), given the first $9$ points of a smooth prototypical trajectory (red).
			Each example resembles a basic motion pattern: constant (a.), accelerated (b.) and curvilinear (c.) motion.}
	\label{fig:simple_pred}
\end{figure}

The pseudo-entropy estimation approach assumes that the learned linear transformation prediction model $\mathcal{M}$ is capable of modeling basic motion patterns in isolation.
In order to verify this, $\mathcal{M}$ has been trained on samples of three clusters corresponding to constant, accelerated and curvilinear motion taken from the decomposed \emph{sdd:hyang04} dataset.
Then, $\mathcal{M}$ was tasked to predict the remainder of each cluster prototype, given its first half ($9$ trajectory points in this case), showing its viability.
The results are depicted in figure \ref{fig:simple_pred}.
The ground truth is depicted in red and the prediction in blue.

\subsection{LVQ Dataset Decomposition: Quality, Consistency and Sensitivity} 
\label{sec:lvq_consistency}
In this section, the viability of the LVQ model given an aligned dataset is evaluated, as it builds the basis of the proposed approach for estimating the information content.
This evaluation employs three exemplary datasets of varying assumed complexity, the \emph{biwi:eth}, \emph{crowds:zara01} and the \emph{sdd:hyang04} dataset.
The evaluation targets the quality and consistency of resulting dataset decompositions, as well as the approach sensitivity to its refinement parameters.
Due to the similarities of datasets used throughout the evaluation, it is assumed that the findings of this section will carry over to the other datasets.

\begin{figure*}[ht] 
	\setlength{\tabcolsep}{15pt}
	\centering
	\begin{tabular}{ccc}
		\includegraphics[width=0.25\textwidth]{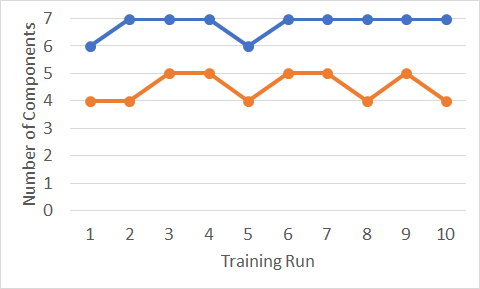} &
		\includegraphics[width=0.25\textwidth]{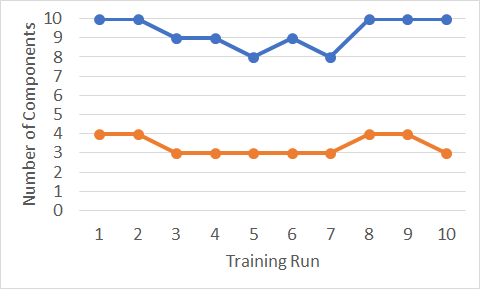} &
		\includegraphics[width=0.25\textwidth]{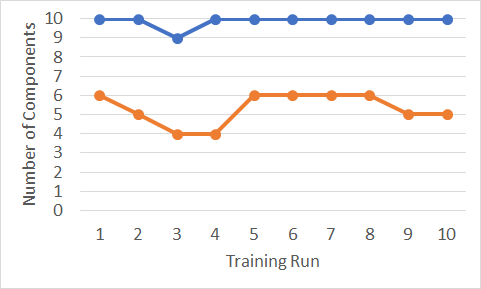} \\
		\includegraphics[width=0.25\textwidth]{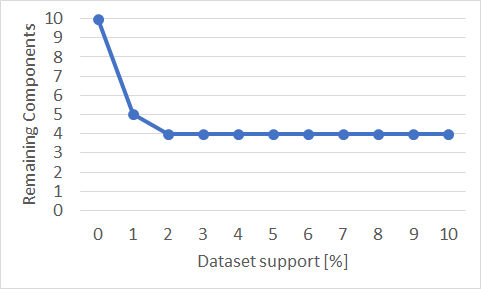} &
		\includegraphics[width=0.25\textwidth]{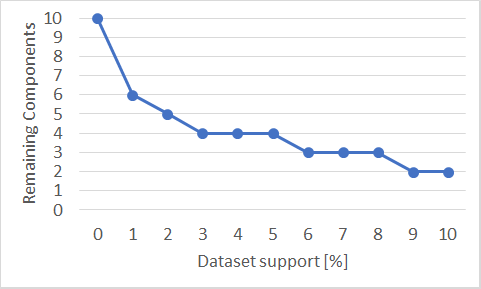} &
		\includegraphics[width=0.25\textwidth]{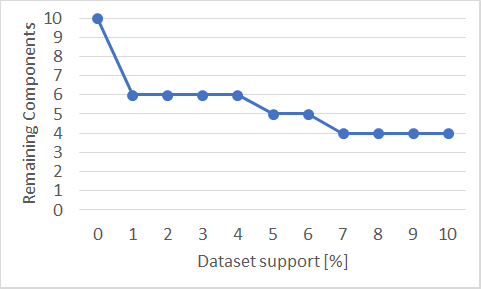} \\
		\includegraphics[width=0.25\textwidth]{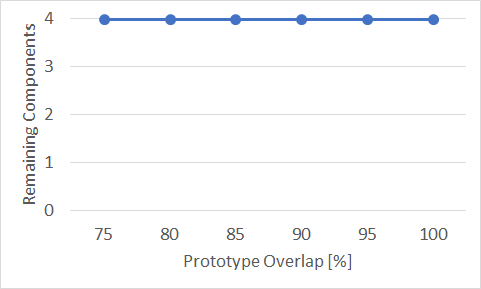} &
		\includegraphics[width=0.25\textwidth]{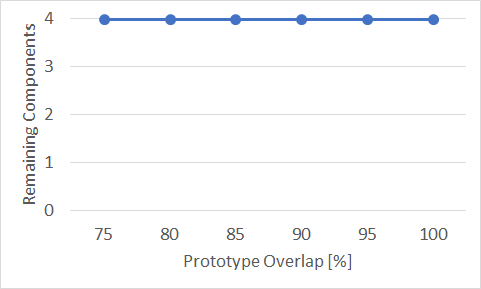} &
		\includegraphics[width=0.25\textwidth]{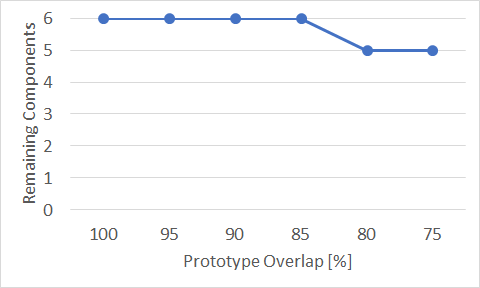} \\
		biwi:eth & crowds:zara01 & sdd:hyang04
	\end{tabular}
	\caption{Number of resulting clusters after multiple training runs before (blue) and after (orange) refinement (1st row) and the influence of different values for the parameters $\epsilon_{\mathrm{phase1}}$ and $\epsilon_{\mathrm{phase2}}$ (2nd and 3rd row) of the LVQ refinement procedure for exemplary datasets.} 
	\label{fig:sensitivity}
\end{figure*}
Starting with consistency, the $10$ available training iterations are examined.
The first row of figure \ref{fig:sensitivity} depicts the number of components identified by the LVQ model before (blue) and after (orange) refinement.
Although there are some small fluctuations, there are no strong deviations which cannot be compensated by averaging.

The influence of the heuristic refinement when varying its parameters $\epsilon_{\mathrm{phase1}}$ and $\epsilon_{\mathrm{phase2}}$ is depicted in the second and third row of figure \ref{fig:sensitivity}.
Looking at the second row, the number of remaining components decreases gradually as expected when increasing $\epsilon_{\mathrm{phase1}}$ from an initial $0\%$ up to $10\%$ necessary support by the data basis.
The third row in figure \ref{fig:sensitivity} indicates the little impact of $\epsilon_{\mathrm{phase2}}$ and phase 2 in general.
Decreasing the minimum number of overlapping trajectory points required for being filtered out from $100\%$ to $75\%$, the final number of components only decreases when there are unnecessary components left, or $\epsilon_{\mathrm{phase2}}$ is too restrictive.
In fact, considering all datasets and training iterations, the second refinement phase only removed at least one component in $185$ of $1823$ cases.

Finally, analyzing the quality of the resulting decomposition, two things have to be considered: the motion patterns represented by each prototype and the significance of each cluster.
As the first point is hard to verify quantitatively, a visual inspection is employed.
Looking at, for example figure \ref{fig:protos_ex} or \ref{fig:res}, the learned prototypes and identified motion patterns appear reasonable for bird's eye view datasets.
This is also discussed briefly in section \ref{ss:discussion}.
For evaluating the quality of the decomposition itself, an approach using a simple prediction model similar to the one described in algorithm \ref{algo:estimate_entropy} can be employed.
While training models on increasingly complex combinations of clusters remains, the test set errors are now compared to the prediction errors on the remaining clusters.
Then, ideally, a significant difference between these errors justifies the existence of the remaining clusters next to the ones combined in the training dataset.
Table \ref{tab:sig_diff_quali} depicts all mean test set and prediction errors (including standard deviations) for clusters generated for \emph{sdd:hyang04}.
Here, all pairwise differences in each row are significant, thus verifying the learned decomposition.
Significance is determined by using a t-test for independent samples using a significance level $\alpha = 0.05$.
Variance homogeneity has been tested using Levene's test and considered when choosing a t-test variant (regular vs. Welch's).
\begin{table}[ht] 
	\centering
	\begin{tabular}{cccccc}
		\textbf{Cluster}     & \textbf{Test Set}& \multicolumn{4}{c}{\textbf{Sub-Dataset Error}} \\
		\textbf{Combination} & \textbf{Error}   & \textbf{2} & \textbf{3} & \textbf{4} & \textbf{5} \\
		\hline
		\multirow{2}{*}{$\mathcal{X}_1$}       & $.0006$ & $.0017$ & $.0017$ & $.0093$ & $.0166$ \\
		& $(.0009)$ & $(.0036)$ & $(.0025)$ & $(.0266)$ & $(.0476)$ \\
		\hdashline
		
		\multirow{2}{*}{$\mathcal{X}_1 \bigcup \mathcal{X}_2$}     & $.0008$ & - & $.0014$ & $.0089$ & $.0157$ \\
		& $(.0027)$ & - & $(.0023)$ & $(.0258)$ & $(.0454)$ \\
		\hdashline
		
		\multirow{2}{*}{$\bigcup^3_{i=1} \mathcal{X}_i$}   & $.0008$ & - & - & $.0090$ & $.0164$ \\
		& $(.0014)$ & - & - & $(.0270)$ & $(.0493)$ \\
		\hdashline
		
		\multirow{2}{*}{$\bigcup^4_{i=1} \mathcal{X}_i$} & $.0013$ & - & - & - & $.0172$ \\
		& $(.0060)$ & - & - & - & $(.0447)$ \\
		\hline
	\end{tabular}
	\caption{Mean test and cluster errors with standard deviations for a simple prediction model trained on different cluster combinations taken from the decomposed \emph{sdd:hyang04}.
			All pairwise differences per row are significant (using $\alpha = 0.05$).
			Note: Errors are unit-less, as the data is in alignment space.}
	\label{tab:sig_diff_quali}
\end{table}

\subsection{Coarse Dataset Ranking}
\label{sec:ranking}
\begin{table*}[ht] 
	\setlength\extrarowheight{2pt}
	\centering
	\begin{tabular}{c|cccccc}
		\textbf{Pseudo-Entropy} & \multicolumn{6}{c}{\textbf{Datasets}} \\
		\hline
		0 & sdd:gates06 & \textbf{sdd:gates07} & sdd:hyang13 & sdd:quad03 && \\
		\hdashline
		\multirow{5}{*}{1} & \textbf{biwi:eth} & biwi:hotel & crowds:zara03 & sdd:bookstore01 & sdd:bookstore04 & sdd:bookstore05 \\
		& sdd:coupa00 & sdd:coupa01 & sdd:coupa03 & sdd:deathcircle02 & sdd:deathcircle04 & sdd:gates00 \\ 
		& sdd:gates02 & sdd:gates05 & sdd:gates08 & sdd:hyang07 & sdd:hyang10 & sdd:hyang14  \\ 
		& sdd:little00 & sdd:little01 & sdd:little02 & sdd:nexus02 & sdd:nexus04 & sdd:nexus08  \\ 
		& sdd:nexus10 &&&&& \\ 
		\hdashline
		\multirow{4}{*}{2} & \textbf{crowds:zara01} & \textbf{crowds:zara02} & sdd:bookstore00 & sdd:bookstore02 & \textbf{sdd:bookstore03} & sdd:bookstore06 \\ 
		& sdd:coupa02 & sdd:gates01 & sdd:gates04 & sdd:hyang01 & sdd:hyang03 & sdd:hyang05  \\ 
		& sdd:hyang06 & sdd:hyang11 & sdd:hyang12 & sdd:nexus00 & sdd:nexus03 & sdd:nexus06  \\ 
		& sdd:nexus07 & sdd:nexus09 & sdd:nexus11 &&& \\
		\hdashline
		\multirow{2}{*}{3} & sdd:deathcircle00 & sdd:deathcircle01 & sdd:gates03 & sdd:hyang00 & sdd:hyang02 & sdd:hyang04 \\ 
		& sdd:little03 & \textbf{sdd:nexus01} &&&& \\
		\hline
	\end{tabular}
	\caption{Coarse complexity ranking of standard trajectory benchmarking datasets based on their estimated information content (pseudo-entropy).
			Higher pseudo-entropy implies higher dataset complexity.
			The datasets are abbreviated in terms of the actual dataset (e.g. \emph{sdd}), the name of the scene included in the dataset (e.g. \emph{hyang}) and the recording number in case there exist multiple recordings for the respective scene.}  
	\label{tab:dataset_ranking}
\end{table*}
Table \ref{tab:dataset_ranking} lists commonly used datasets, grouped by their average rounded pseudo-entropy values, calculated according to algorithm \ref{algo:estimate_entropy} for each trained pair of alignment and LVQ models.
Again, a t-test for independent samples with a significance level of $\alpha = 0.05$ is used for testing significance, using Levene's test for testing variance homogeneity.
Some scenes were filtered out after the setup, as their number of samples did not exceed $150$, leading to less stable alignment and LVQ models.
The evaluation resulted in a coarse ranking of datasets, with datasets being assigned to one of four groups of similar pseudo-entropy.
It can be seen that the \emph{biwi} scenes are among the datasets containing the least information, while the most informative scenes are found in the \emph{sdd} dataset.
For demonstration purposes, the datasets and prototypes for \emph{biwi:eth} and \emph{sdd:nexus01} are illustrated in figure \ref{fig:res} for similar sequence lengths ($12$ and $15$).
As opposed to \emph{biwi:eth}, \emph{sdd:nexus01}, being ranked has containing more information, consists of three times the amount of motion patterns, including constant velocity, curvilinear motion, acceleration and deceleration, as well as a mixed motion pattern.
This mixed pattern consists of constant velocity, decelerating and accelerating parts, and might occur due to the rather high sequence length with respect to the covered time span.

\subsection{Discussion}
\label{ss:discussion}
Conclusively, selected aspects of the proposed approach and employed methodology are discussed. 
Then, potential factors for creating a more fine-grained ranking and some insights gained from the analyses are discussed.

\subsubsection{Approach and Methodology}
In the context of dataset complexity assessment, normalizing the velocities using the alignment model should be discussed.
Given a prototypical motion pattern, variations in velocity can be generated by scaling it accordingly, thus it is assumed that the pattern itself is the main contributor to a higher dataset entropy.
In case the original motion patterns are required, recall that the clustering pipeline corresponds to the encoding part of a well-established auto-encoding architecture, thus the original velocities can be recovered from the dataset representation when employing the full auto-encoding architecture.

Further, a dataset-dependent trajectory length $M$ was chosen instead of a common value for all datasets.
This decision is motivated by large differences in trajectory lengths occurring in existing datasets as well as unknown ground resolutions.
While differences in length may only lead to some datasets becoming very small if $M$ is too high, it is not clear if two trajectories of the same length from different datasets are comparable due to unknown ground resolutions.
Even if the average offset between subsequent trajectory points is equal, both trajectories might represent different real-world speeds, thus covering a longer/shorter distance in reality, which directly impacts occurring motion patterns and the influence of agent-agent interaction.

The pseudo-entropy aims to reveal the average amount of information contained in a given dataset.
Looking at the coarse ranking in section \ref{sec:ranking}, the results appear reasonable from an experience point of view.
For verifying this coarse ranking in an experimental setup using state-of-the-art trajectory prediction models, it would be necessary to put all datasets in a common reference frame and re-sample trajectories to achieve a matching ground resolution, i.e. the distance between subsequent trajectory points of objects moving at the same real-world speed must be equal, for all datasets.
This can be an interesting experiment for future work on the topic of dataset complexity analysis.

Lastly, the presented approach only allows for a coarse complexity ranking of given datasets.
For achieving a more fine-grained ranking, additional factors need to be considered.
Possible factors are discussed next.

\begin{figure*}[ht]
	\setlength{\tabcolsep}{3pt}
	\centering
	\begin{tabular}{cccccc}
		\includegraphics[width=0.15\textwidth]{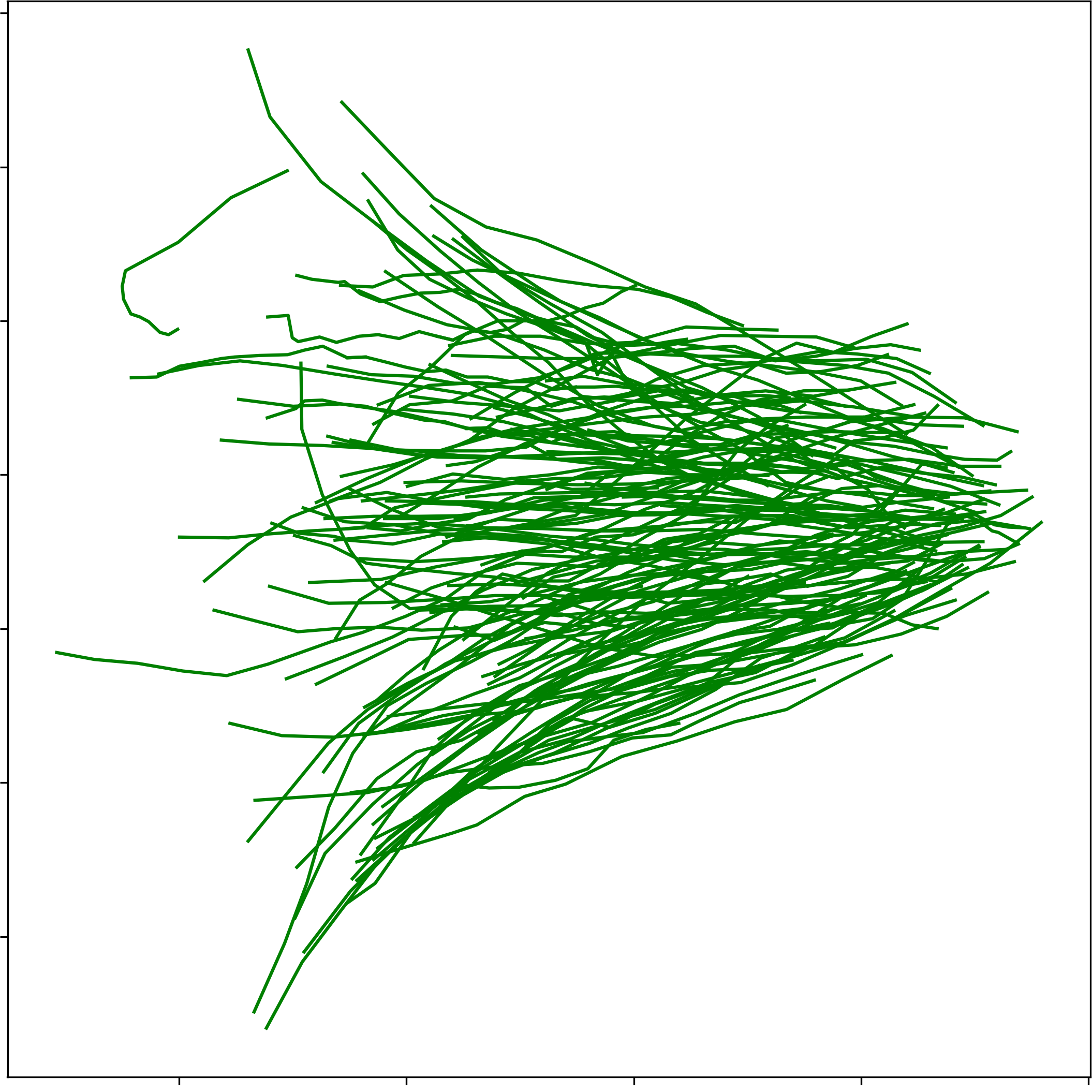} &
		\includegraphics[width=0.15\textwidth]{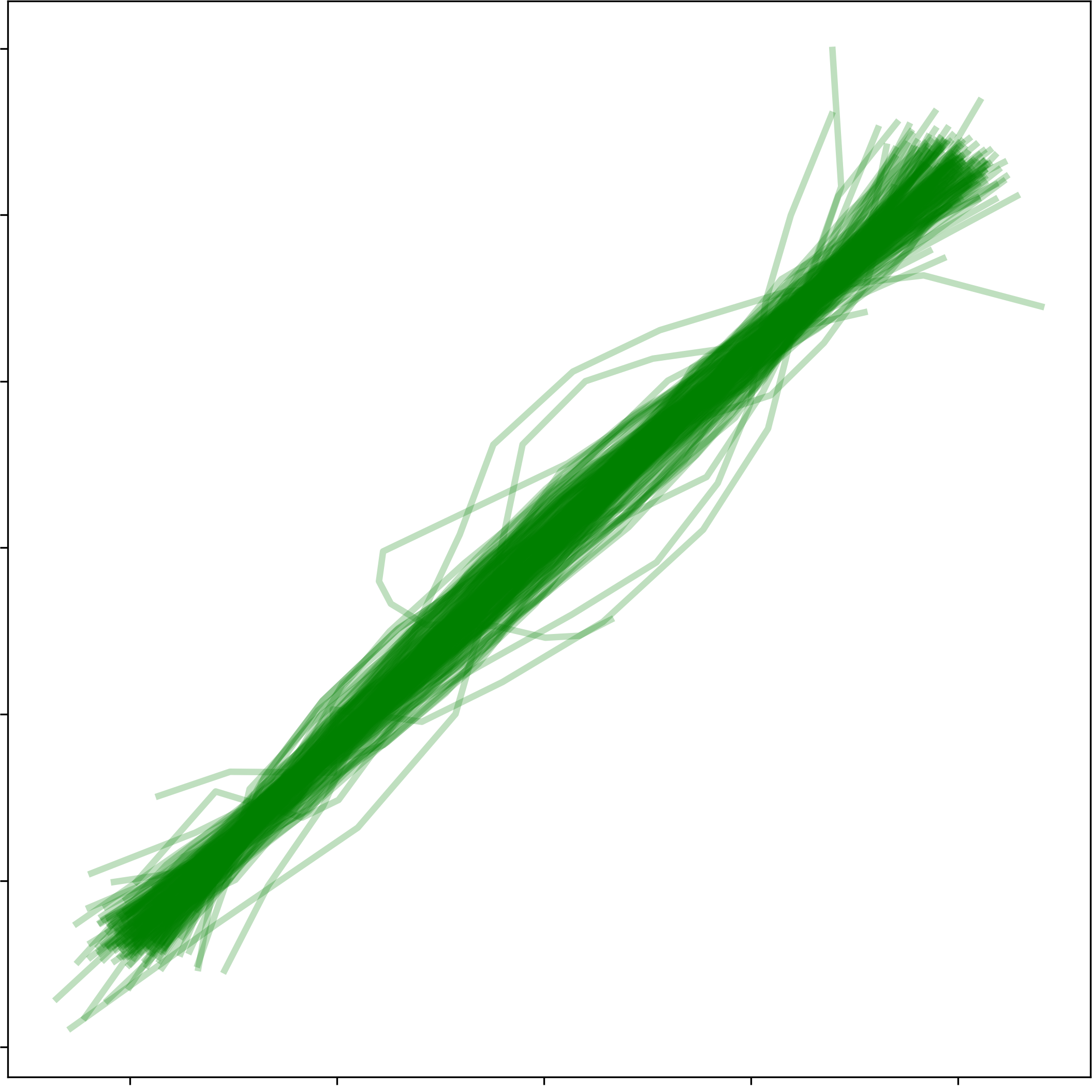} &
		\includegraphics[width=0.15\textwidth]{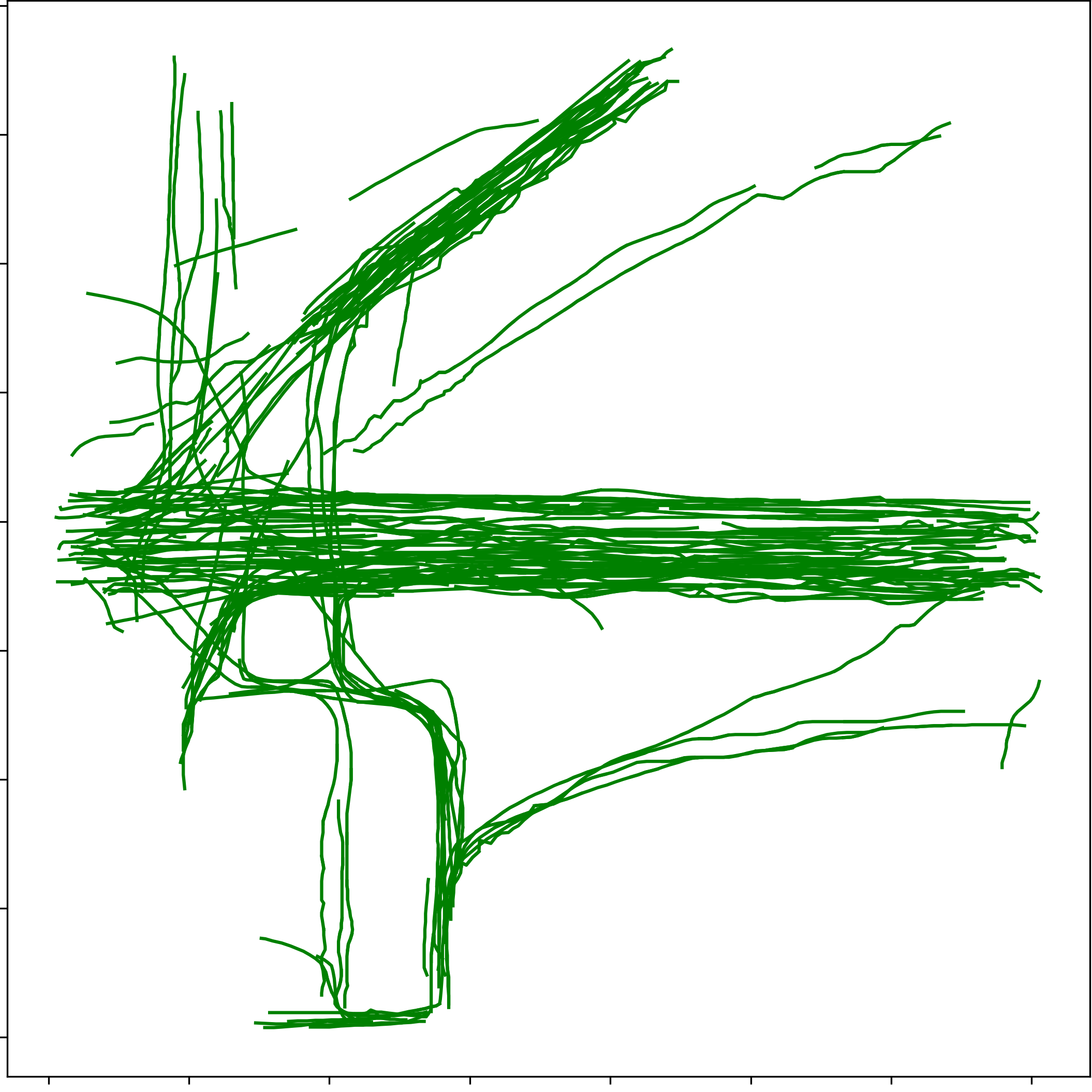} &
		\includegraphics[width=0.15\textwidth]{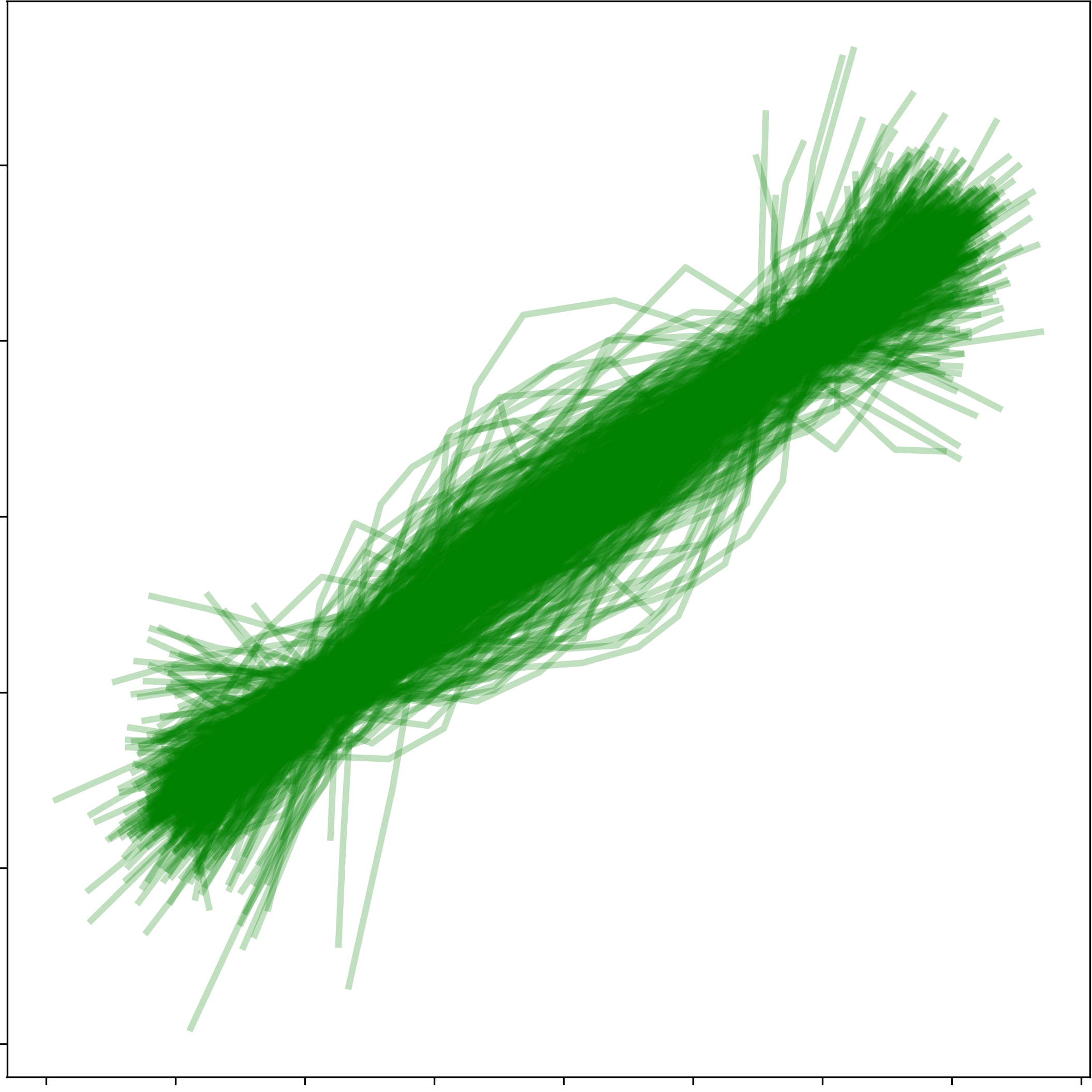} &
		\includegraphics[width=0.15\textwidth]{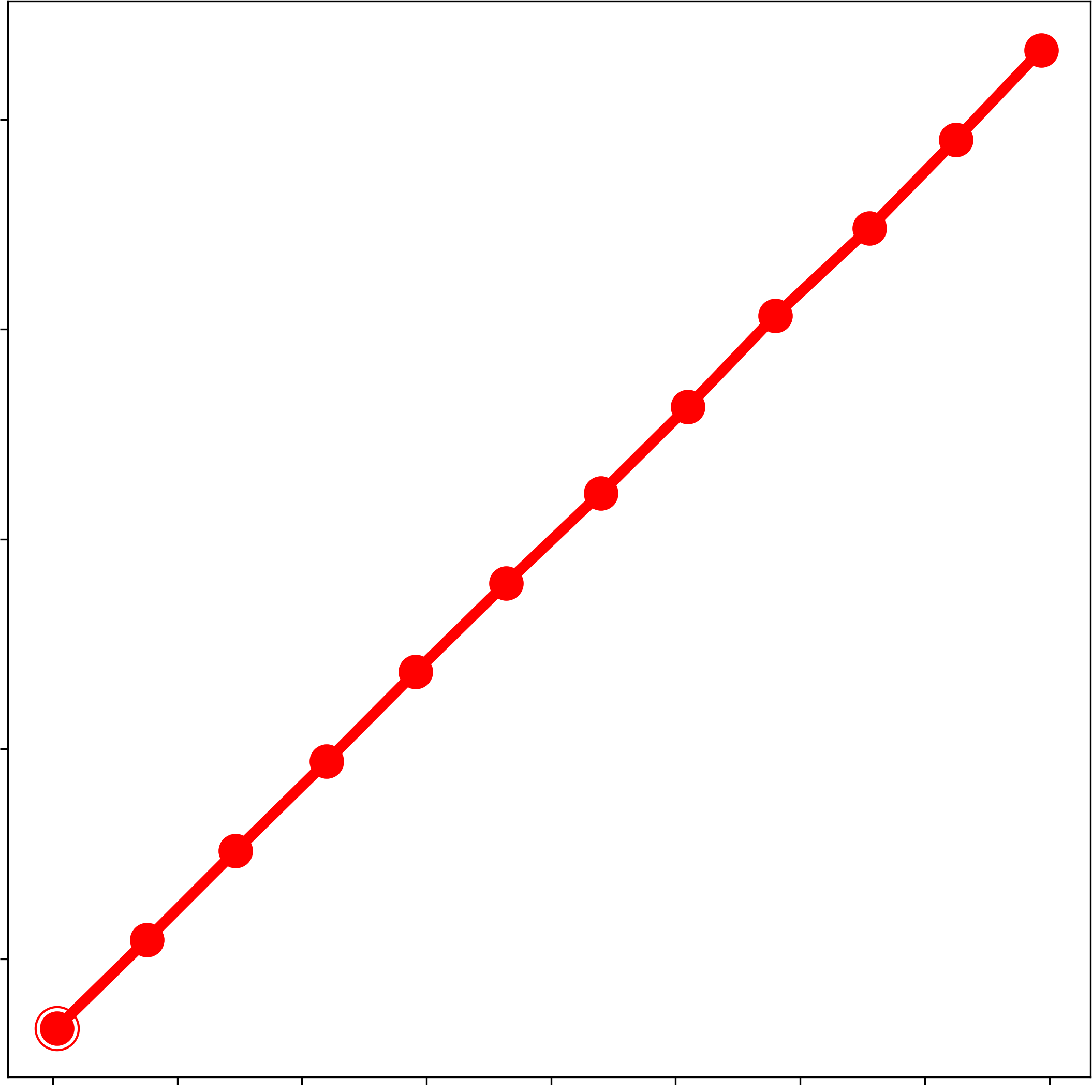} &
		\includegraphics[width=0.15\textwidth]{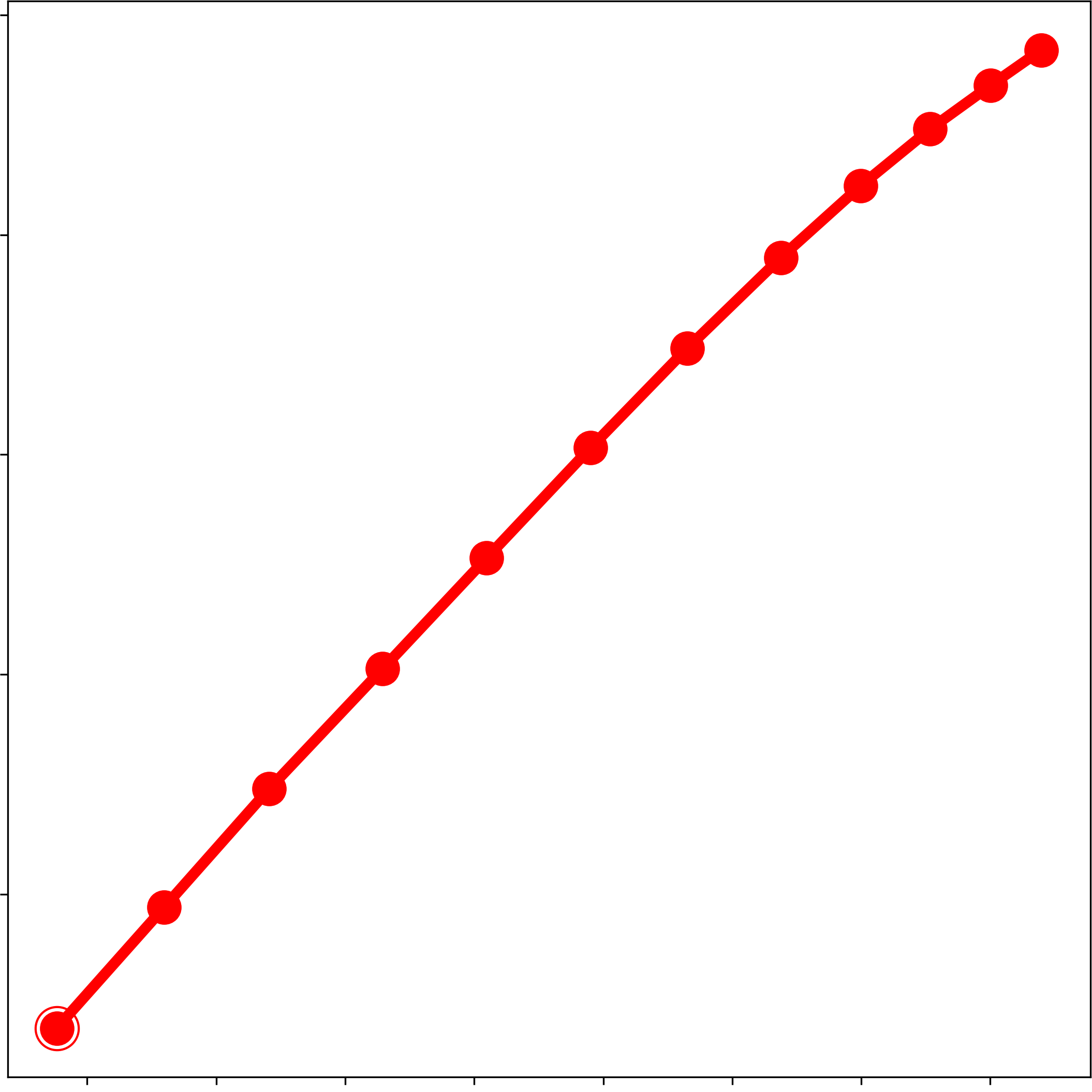} \\
		a. & b. & c. & d. & e. & f. \\
		
		\includegraphics[width=0.15\textwidth]{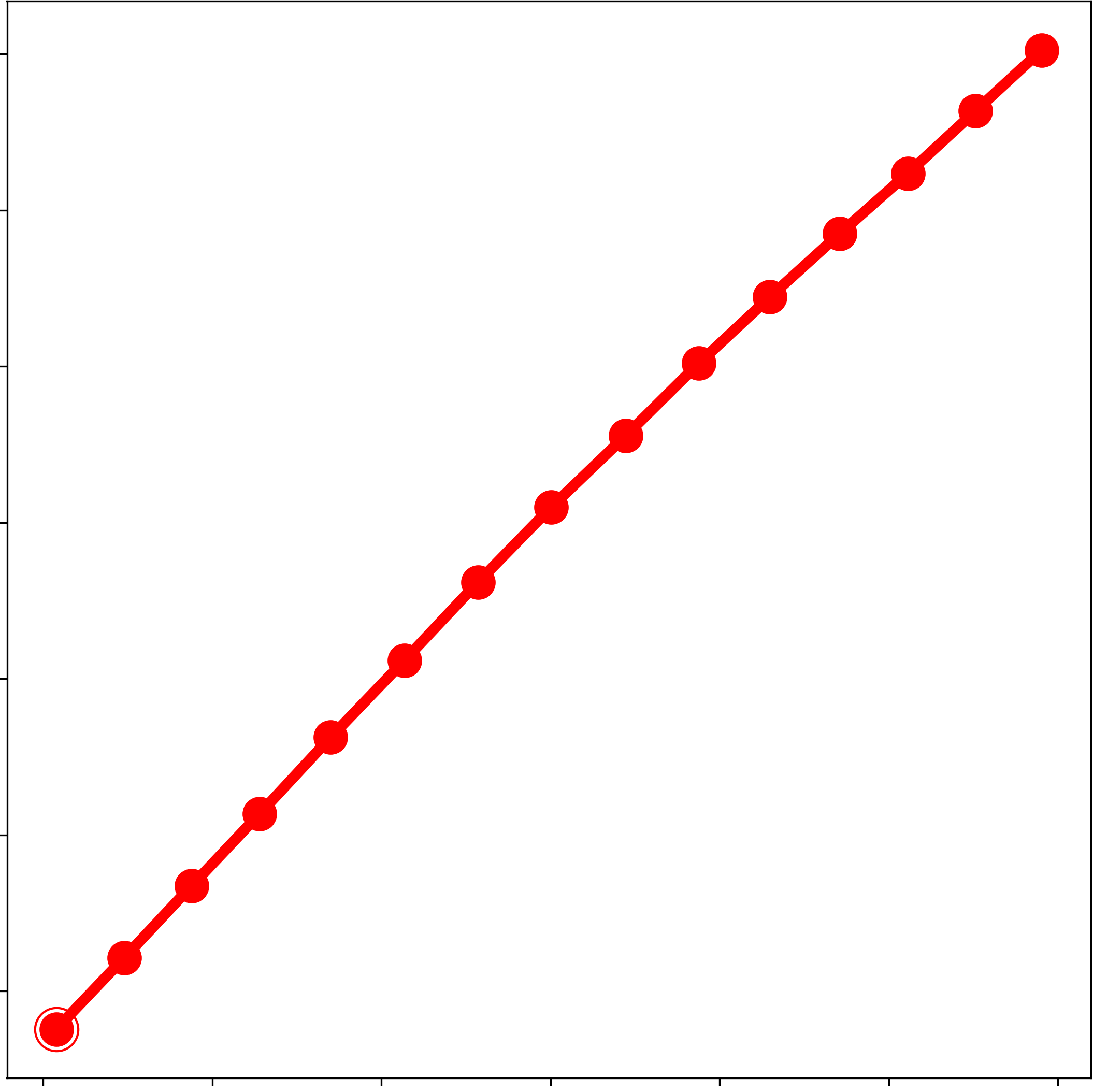} &
		\includegraphics[width=0.15\textwidth]{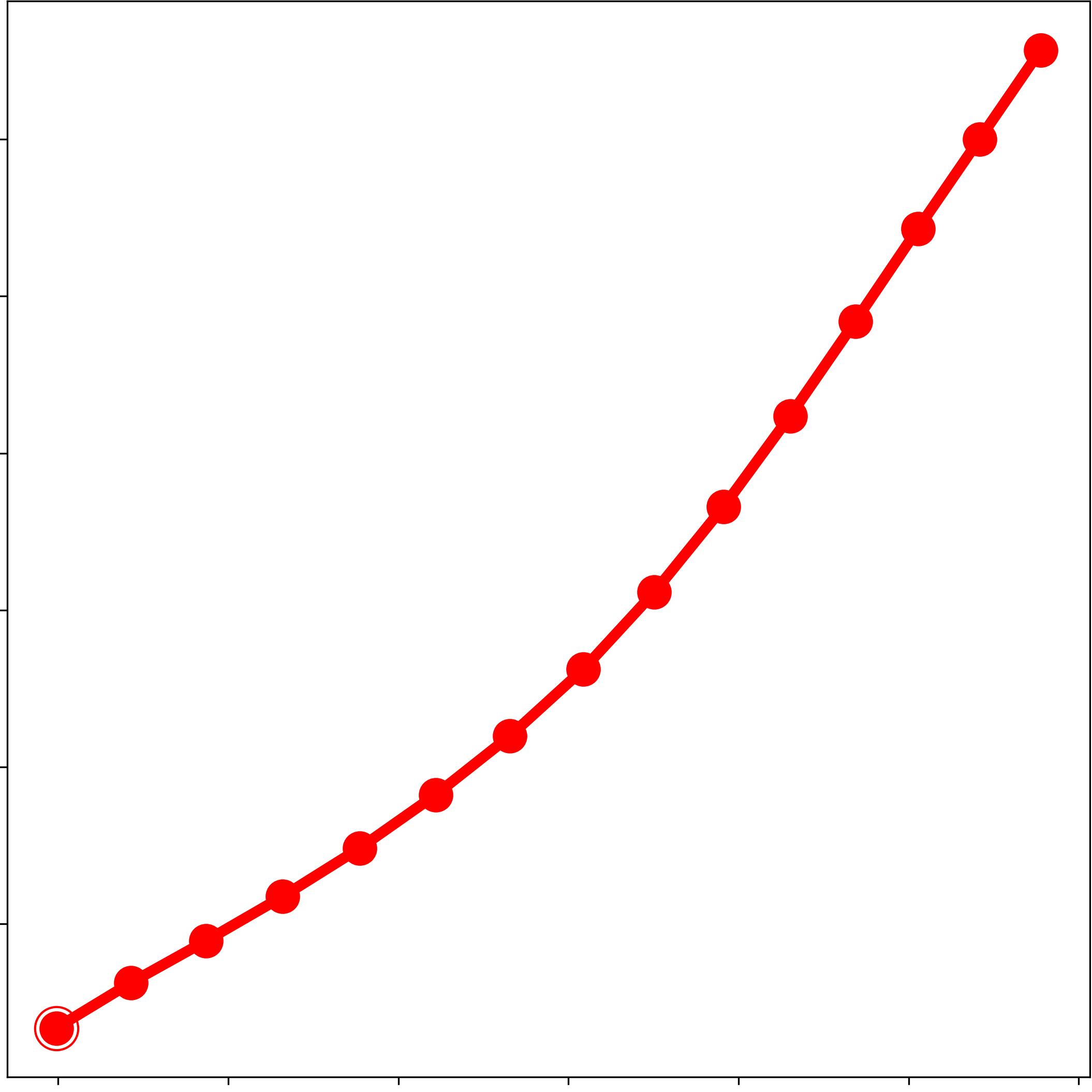} &
		\includegraphics[width=0.15\textwidth]{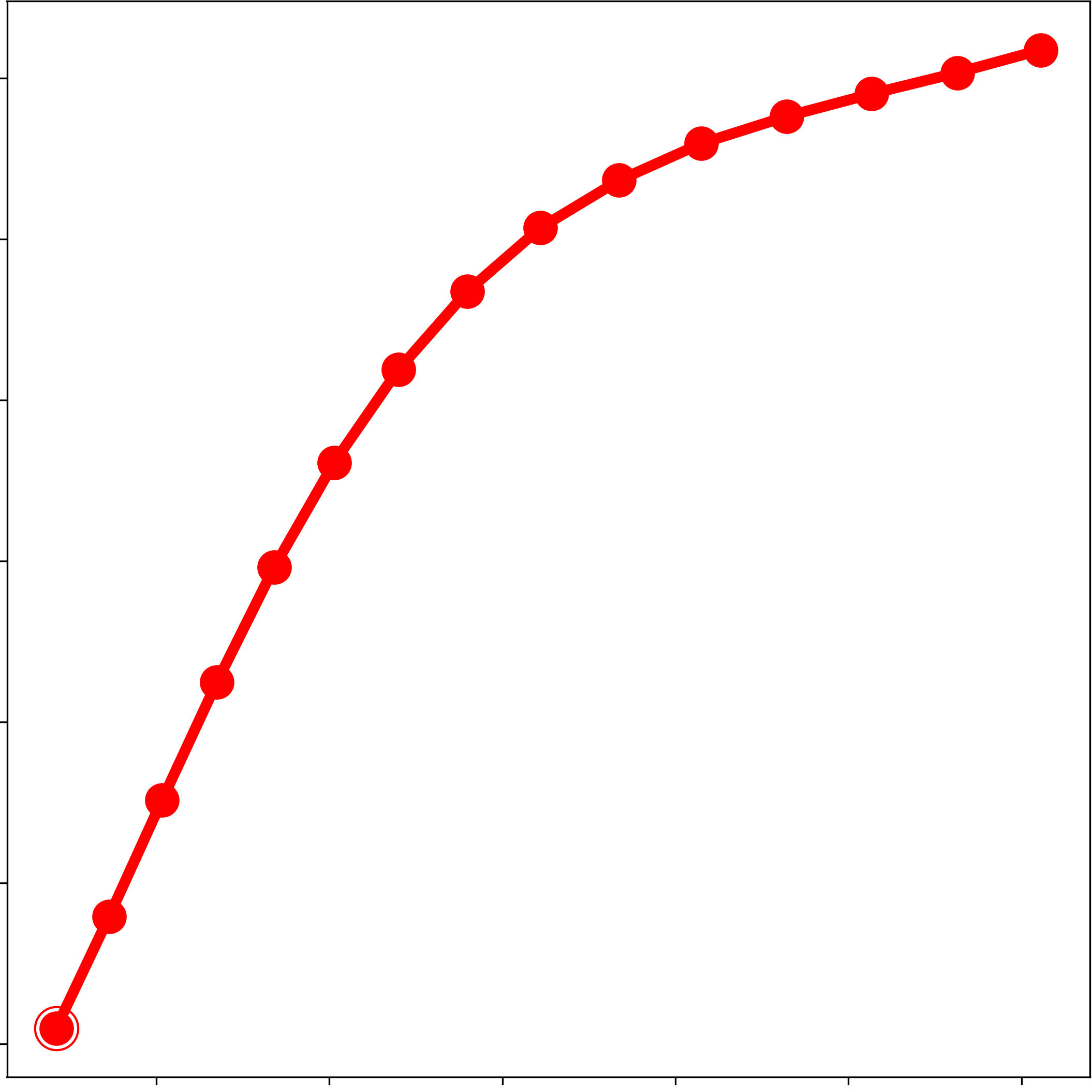} &
		\includegraphics[width=0.15\textwidth]{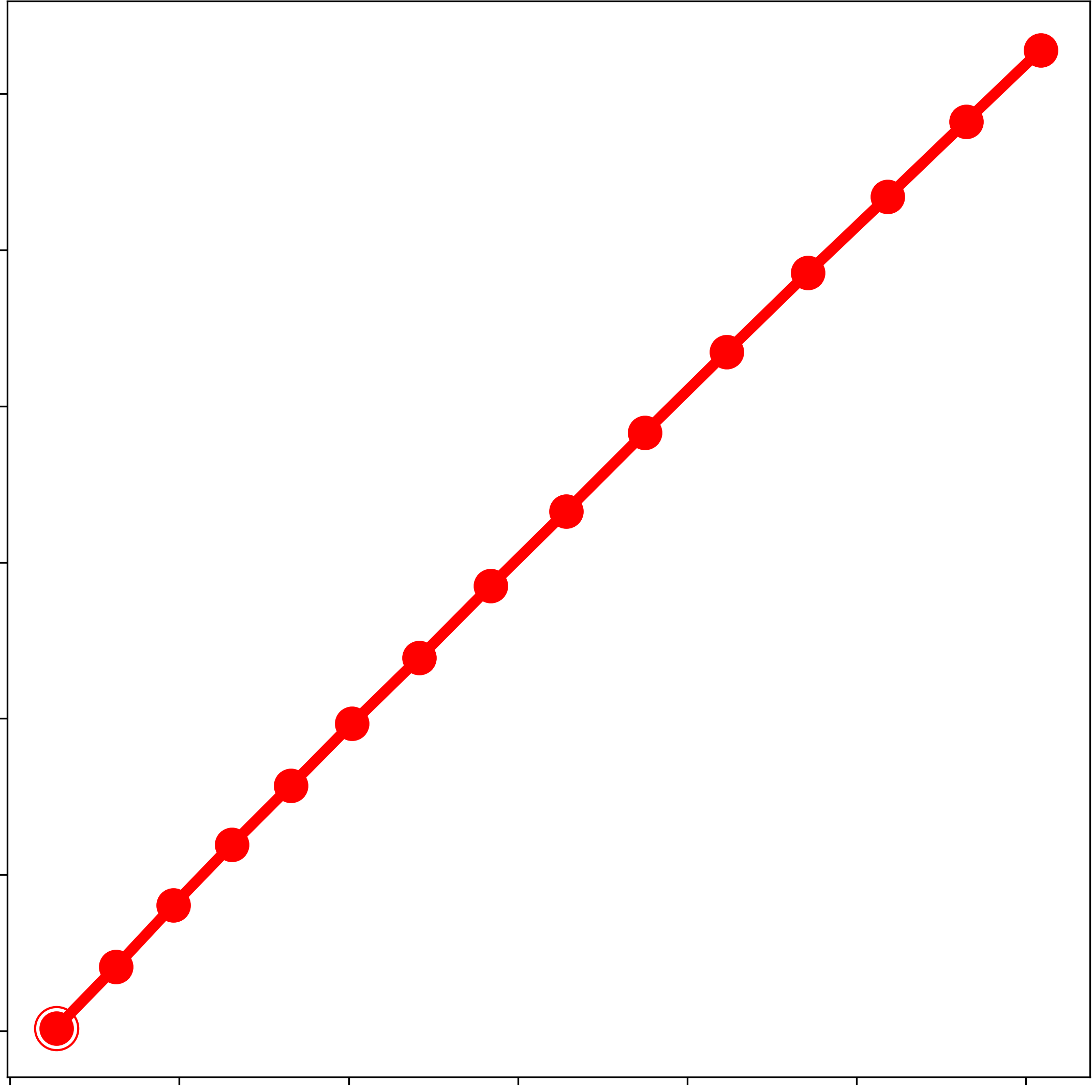} &
		\includegraphics[width=0.15\textwidth]{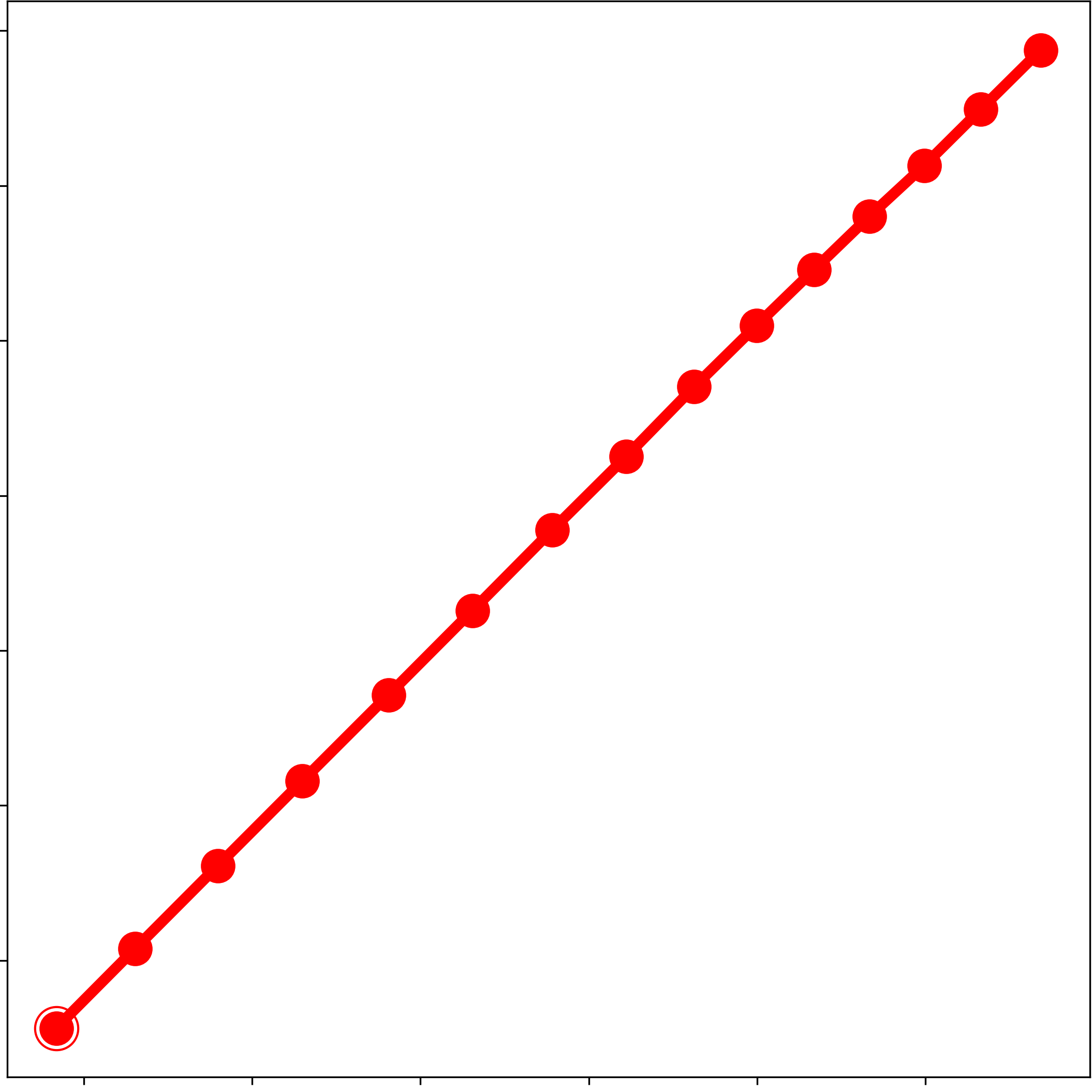} & 
		\includegraphics[width=0.15\textwidth]{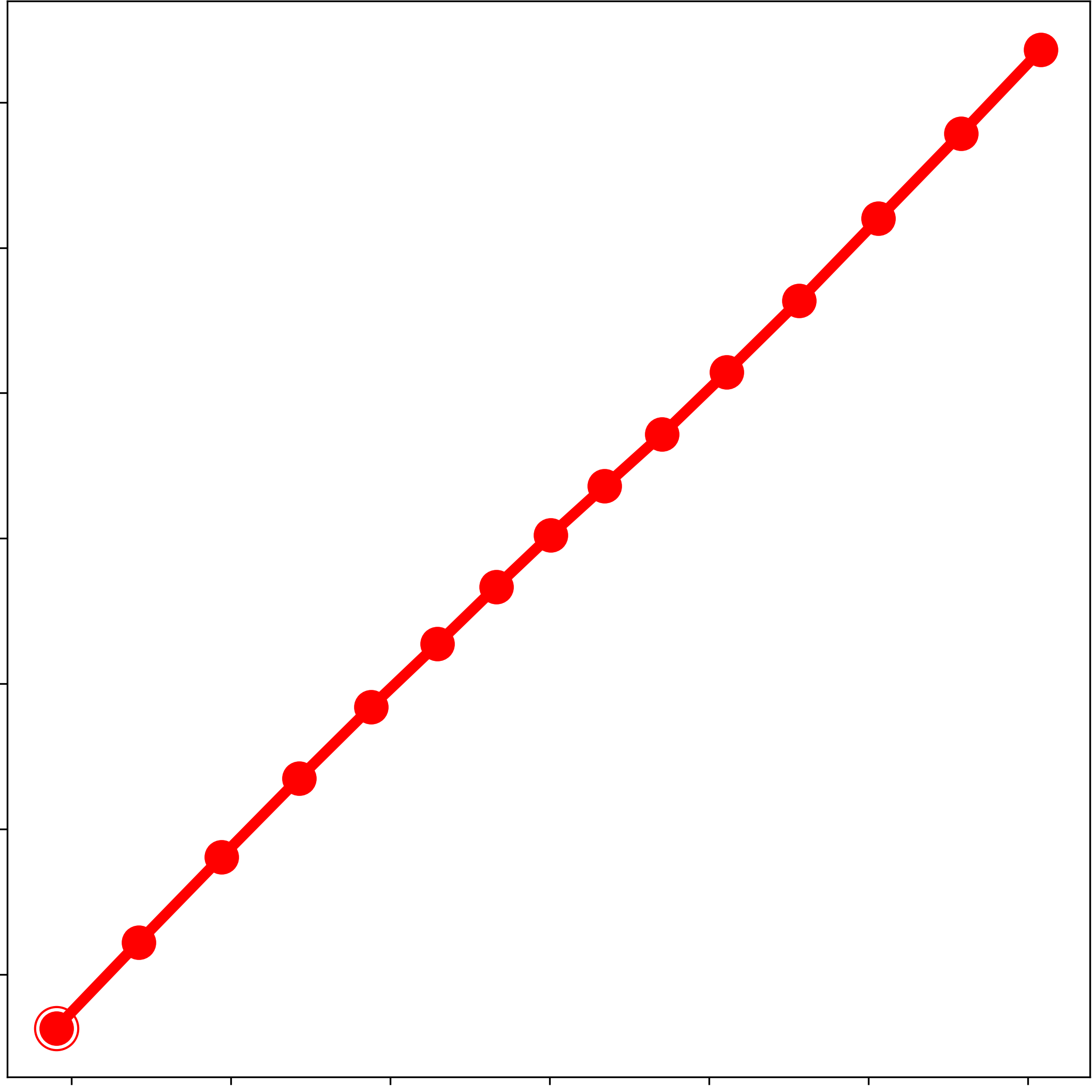} \\
		g. & h. & i. & j. & k. & l.
	\end{tabular}
	\caption{Data and aligned data (a. \& b.: \emph{biwi:eth}, c. \& d: \emph{sdd:nexus01}) with similar sequence lengths ($12$ \& $15$), as well as learned prototypes for \emph{biwi:eth} (e. \& f.) and \emph{sdd:nexus01} (g. -- l.).
		Having a higher complexity score, \emph{sdd:nexus01} consists of a higher variety of motion patterns, including constant velocity (g.), curvilinear motion (h. \& i.), acceleration (j.), deceleration (k.) and a mixed pattern (l.).
		The fraction of samples assigned to each prototype are $93\%$ (a.) and $7\%$ (b.) for \emph{biwi:eth} and $43.5\%$ (g.), $19.4\%$ (h.), $13.7\%$ (i.), $8.6\%$ (j.), $5.6\%$ (k.) and $9.2\%$ (l.) for \emph{sdd:nexus01}.} 
	\label{fig:res}
\end{figure*}

\subsubsection{Potential Factors Affecting Complexity}
\label{ss:factors}
Multiple factors contributing to dataset complexity could be derived from a learned dataset decomposition.
First, the diversity between motion patterns, covered implicitly in the pseudo-entropy, could be considered explicitly.
In case of distinguishable patterns, statistical models need to be capable of capturing multiple modes in the data, requiring a higher modeling capacity.
Thus, a higher pattern diversity is expected to correspond to a higher dataset complexity.
The second factor considers occurring variations of the same pattern.
This factor looks promising, as a higher variation implies a higher uncertainty when modeling specific motion patterns, making it harder to capture by using statistical models.
Lastly, the relevance distribution of identified motion pattern could be considered.
This mainly focuses on biases in the data, and thus answers the question if there is a prevalent motion pattern or if the occurrence of all patterns is equally likely.
Then, less biased datasets can be considered as more complex, as less biased data enables statistical models to capture different patterns in the first place.

Beyond that, agent-agent interaction, as well as the environmental cues can play an integral role in assessing trajectory dataset complexity.
Looking at interaction, its influence on the shape of single trajectories can be significant, though this heavily depends on the chosen sample frequency as well as the ground resolution of a given dataset.
More specifically, the influence of agent-agent interaction becomes less relevant, the sparser a trajectory is sampled, due to interactions mainly occurring on short time scales.
The same applies to the ground resolution, where interactions become less visible when the spatial distance between subsequent trajectory points increases.
As a final note, sensor noise must be considered, as there is a risk of interactions being indistinguishable from noise.
For environmental cues, positional biases caused for example by junctions can heavily impact the occurrence of specific motion patterns, especially curvilinear motion.
This leads to more diverse, and thus to potentially more complex datasets.

\subsubsection{Interesting Findings}
All datasets in this comparison are recorded from a birds-eye view.
Inherent to this perspective is the expectation that there are common motion patterns in all datasets, independent of the time horizon.
This fact has been, with a few exceptions, confirmed, in that almost all scenes contain slight variations of at least one basic motion pattern, including constant, accelerated, decelerated and curvilinear motion. 
Some datasets contain multiple variations of the same basic motion pattern or even mixed motion patterns, enabled by the, partially, high sequence length $M$. 
This can be seen in figure \ref{fig:res} (g. -- l.).

\begin{figure}[h] 
	\centering
	\includegraphics[width=0.95\columnwidth]{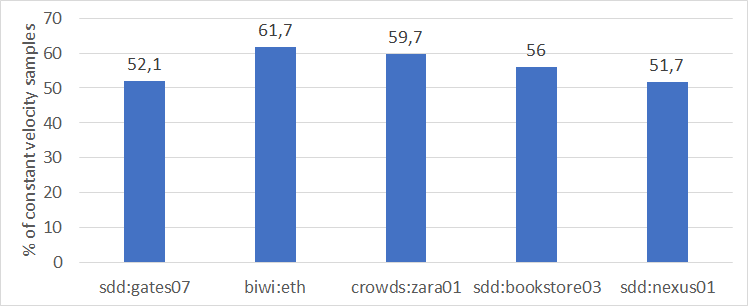}
	\caption{Depicts the fraction of samples assigned to the constant velocity prototype (sample ratio) for a range of low to high complexity scene datasets.
		The fractions are calculated by taking the mean sample ratio, taking multiple iterations and sequence lengths into account.
		Even with a high number of distinct motion patterns in the data (e.g. \emph{sdd:nexus01}), constant velocity samples make up $52\%$ of all samples in the dataset.}
	\label{fig:cv_chart} 
\end{figure}
Another aspect related to the motion patterns found in the data is, that in all datasets, the constant velocity pattern is the dominant, i.e. most supported, motion pattern, covering a large fraction of the entire dataset (see figure \ref{fig:cv_chart} for exemplary fractions for low to high complexity scene datasets).
This has multiple implications related to common evaluation methodology in current state-of-the-art publications.
On the one hand, it is a perfect explanation for the difficulties in beating a simple linear extrapolation model in the task of human trajectory prediction.
This phenomenon could for example be observed during the TrajNet challenge \cite{sadeghiankosaraju2018trajnet}, where multiple of the first submissions failed to beat the linear model.
On the other hand, this fact indicates, that an arbitrarily assembled benchmarking data basis poses the risk of rendering corner cases, i.e. motion patterns different from the constant velocity pattern, statistically irrelevant.
This leads to statistical models that are incapable of modeling more complex motion and also struggle with beating a linear model.

\subsubsection{Comparison with OpenTraj}
Dataset complexity estimation in the context of human trajectory prediction still poses an unexplored topic in the literature.
Because of that, there is no quantitative approach for measuring and comparing the performance of different approaches for dataset complexity estimation.
As a result, this section resorts to a qualitative comparison of the pseudo-entropy approach presented in this paper and the OpenTraj approach and aims to serve as a verification of results.
Using the result provided in \cite{amirian2020opentraj}, a superficial comparison can be made by comparing the coarse dataset ranking based on pseudo-entropy with the clustering and entropy analyses in OpenTraj (figure 3 in \cite{amirian2020opentraj}).
It has to be noted, that in this paper every scene of each dataset has been treated independently (e.g. sdd:deathcircle - scene 1), while OpenTraj pooled the results for each individual dataset.
Being common to both evaluation sections, the \emph{eth hotel}, \emph{zara} and \emph{sdd:deathcircle} datasets can be used as a sample for the comparison.
Looking at the pseudo-entropy based coarse ranking, these datasets provide a dataset of low, medium and high entropy, respectively.
This relative ranking is consistent with the findings provided by OpenTraj, confirming the plausibility of both approaches.

\section{Concluding Remarks} 
\label{sec:summary}
In the context of statistical learning, dataset complexity is closely related to the entropy of a given dataset.
Thus, an approach for estimating the amount of information contained in trajectory datasets was proposed.
The approach relies on a velocity-agnostic dataset representation generated by an alignment followed by vector quantization.  
Using this approach, a coarse complexity ranking of commonly used benchmarking datasets has been generated.
A following discussion addressed the results and methods used, as well as interesting findings based on the analyses, stressing the importance of a well-rounded data basis.

\subsubsection{Implications for the State of Benchmarking} 
The approach, methods and results presented in this paper can be valuable in the context of dataset and prediction model analysis, as well as benchmarking in general.
First of all, the spatial sequence alignment combined with the LVQ approach can be used for analyzing datasets on different timescales, e.g. for extracting underlying motion patterns.
This analysis especially benefits the selection of observation and prediction sequence lengths in benchmarking, as well as the selection of an appropriate prediction model in terms of model capacity.
The latter is motivated by the fact, that low-capacity models usually suffice for less complex data, which might in turn reduce cases of over-fitting and unnecessary computational effort.
Further, the resulting dataset decomposition can be used to enhance qualitative analyses of prediction model capabilities in cases where the model might struggle with specific subsets of the data.
Lastly, a hierarchy of tasks within a benchmark with increasing difficulty could be built on the dataset decomposition in combination with the presented coarse dataset ranking.

\subsubsection{Future Research Direction}
This paper aims to constitute a step towards thorough dataset complexity analysis.
The following paragraphs try to give some open research directions in order to expand on the approach and findings of this paper.

\textbf{Consider Model Uncertainty.} 
Currently, model uncertainty is only considered implicitly by averaging multiple instances of the presented pipeline when estimating the dataset pseudo-entropy.
However, the variance of the ensemble is disregarded in this proof of concept.
Thus it could be interesting to examine if explicitly incorporating the models' uncertainty about its output could benefit the entropy estimation.

\textbf{Birds-eye View.} 
So far, all compared datasets are birds-eye view datasets.
While this view is the common case\footnote{A common alternative to birds-eye view datasets are datasets with trajectories 3D-projected onto a flat ground plane. The resulting datasets are comparable to birds-eye view datasets and usually only differ in scale.} for long-term human trajectory prediction datasets, trajectory complexity analysis is also relevant for other views (e.g. frontal view).
Considering the structure of the presented approach, the entropy estimation should work as long as the spatial sequence alignment is applicable to the scenario of interest.
In the current state, the alignment model expects complete tracklets as input and thus does not have to cope with missing observations, for example arising through occlusions when using a frontal view.
Following this, the alignment model should be extended accordingly and be evaluated on a range of different datasets with varying views and object types.

\appendices
\setcounter{table}{0}
\renewcommand\thetable{\Alph{section}.\arabic{table}}

\section{Dataset Details}
\label{app:datasets}
In order to give more details on the datasets used throughout this paper, table \ref{tab:dataset_details} lists the number of samples included in each dataset, the recording conditions (location and acquisition) as well the the average trajectory length (with standard deviation).
In accordance with the evaluation section \ref{sec:evaluation}, the annotation rate has been aligned for all datasets to a fixed rate of $2.5$ annotations per second.
\begin{table*}[htb]
	\centering	
	\begin{tabular}{lcccc}
		\hline
		\textbf{Dataset} & \textbf{\# Samples} & \textbf{Avg. Trajectory Length ($\mu \pm \sigma$)} & \textbf{Location} & \textbf{Acquisition} \\
		\hline
		biwi:eth & $354$ & $15.48 \pm 8.01$ & university entrance & top view camera \\
		biwi:hotel & $378$ & $17.25 \pm 12.14$ & urban street & top view camera \\
		crowds:zara01 & $148$ & $34.82 \pm 16.61$ & urban street & top view camera \\
		crowds:zara02 & $204$ & $47.66 \pm 72.31$ & urban street & top view camera \\
		crowds:zara03 & $137$ & $36.53 \pm 32.81$ & urban street & top view camera \\
		sdd:bookstore00 & $198$ & $84.29 \pm 140.11$ & university campus & drone camera \\
		sdd:bookstore01 & $266$ & $50.73 \pm 71.27$ & university campus & drone camera \\
		sdd:bookstore02 & $226$ & $49.38 \pm 73.31$ & university campus & drone camera \\
		sdd:bookstore03 & $254$ & $43.34 \pm 40.10$ & university campus & drone camera \\
		sdd:bookstore04 & $138$ & $29.86 \pm 26.43$ & university campus & drone camera \\
		sdd:bookstore05 & $142$ & $30.56 \pm 31.62$ & university campus & drone camera \\
		sdd:bookstore06 & $112$ & $38.20 \pm 34.39$ & university campus & drone camera \\
		sdd:coupa00 & $105$ & $81.12 \pm 117.44$ & university campus & drone camera \\
		sdd:coupa01 & $75$ & $66.92 \pm 100.15$ & university campus & drone camera \\
		sdd:coupa02 & $86$ & $54.09 \pm 69.30$ & university campus & drone camera \\
		sdd:coupa03 & $105$ & $112.56 \pm 186.14$ & university campus & drone camera \\
		sdd:deathCircle00 & $579$ & $45.36 \pm 32.58$ & university campus & drone camera \\
		sdd:deathCircle01 & $856$ & $35.81 \pm 28.44$ & university campus & drone camera \\
		sdd:deathCircle02 & $29$ & $24.90 \pm 10.89$ & university campus & drone camera \\
		sdd:deathCircle03 & $712$ & $33.36 \pm 21.98$ & university campus & drone camera \\
		sdd:deathCircle04 & $36$ & $21.33 \pm 11.76$ & university campus & drone camera \\
		sdd:gates00 & $112$ & $33.30 \pm 23.55$ & university campus & drone camera \\
		sdd:gates01 & $255$ & $39.67 \pm 33.65$ & university campus & drone camera \\
		sdd:gates02 & $114$ & $34.83 \pm 36.36$ & university campus & drone camera \\
		sdd:gates03 & $341$ & $38.24 \pm 29.18$ & university campus & drone camera \\
		sdd:gates04 & $81$ & $35.16 \pm 22.50$ & university campus & drone camera \\
		sdd:gates05 & $38$ & $35.74 \pm 24.78$ & university campus & drone camera \\
		sdd:gates06 & $17$ & $26.71 \pm 26.05$ & university campus & drone camera \\
		sdd:gates07 & $33$ & $26.85 \pm 18.55$ & university campus & drone camera \\
		sdd:gates08 & $66$ & $37.39 \pm 34.06$ & university campus & drone camera \\
		sdd:hyang00 & $283$ & $60.81 \pm 44.88$ & university campus & drone camera \\
		sdd:hyang01 & $126$ & $73.51 \pm 74.42$ & university campus & drone camera \\
		sdd:hyang02 & $171$ & $73.11 \pm 122.55$ & university campus & drone camera \\
		sdd:hyang03 & $178$ & $51.66 \pm 40.23$ & university campus & drone camera \\
		sdd:hyang04 & $356$ & $71.35 \pm 51.94$ & university campus & drone camera \\
		sdd:hyang05 & $124$ & $72.11 \pm 109.38$ & university campus & drone camera \\
		sdd:hyang06 & $108$ & $68.80 \pm 59.96$ & university campus & drone camera \\
		sdd:hyang07 & $35$ & $33.20 \pm 16.03$ & university campus & drone camera \\
		sdd:hyang08 & $9$ & $34.11 \pm 14.49$ & university campus & drone camera \\
		sdd:hyang09 & $7$ & $11.86 \pm 8.20$ & university campus & drone camera \\
		sdd:hyang10 & $60$ & $60.68 \pm 106.97$ & university campus & drone camera \\
		sdd:hyang11 & $138$ & $48.14 \pm 41.75$ & university campus & drone camera \\
		sdd:hyang12 & $64$ & $39.28 \pm 31.93$ & university campus & drone camera \\
		sdd:hyang13 & $32$ & $22.94 \pm 19.93$ & university campus & drone camera \\
		sdd:hyang14 & $22$ & $52.59 \pm 36.56$ & university campus & drone camera \\
		sdd:little00 & $51$ & $31.94 \pm 24.58$ & university campus & drone camera \\
		sdd:little01 & $139$ & $26.83 \pm 26.13$ & university campus & drone camera \\
		sdd:little02 & $95$ & $20.11 \pm 13.44$ & university campus & drone camera \\
		sdd:little03 & $210$ & $45.81 \pm 44.19$ & university campus & drone camera \\
		sdd:nexus00 & $113$ & $65.64 \pm 34.56$ & university campus & drone camera \\
		sdd:nexus01 & $176$ & $68.30 \pm 44.18$ & university campus & drone camera \\
		sdd:nexus02 & $150$ & $50.07 \pm 36.08$ & university campus & drone camera \\
		sdd:nexus03 & $32$ & $47.12 \pm 28.32$ & university campus & drone camera \\
		sdd:nexus04 & $43$ & $38.16 \pm 26.89$ & university campus & drone camera \\
		sdd:nexus05 & $12$ & $30.50 \pm 10.75$ & university campus & drone camera \\
		sdd:nexus06 & $90$ & $63.10 \pm 37.16$ & university campus & drone camera \\
		sdd:nexus07 & $115$ & $57.19 \pm 36.30$ & university campus & drone camera \\
		sdd:nexus08 & $104$ & $50.16 \pm 32.84$ & university campus & drone camera \\
		sdd:nexus09 & $89$ & $69.02 \pm 60.36$ & university campus & drone camera \\
		sdd:nexus10 & $55$ & $52.38 \pm 36.31$ & university campus & drone camera \\
		sdd:nexus11 & $49$ & $61.22 \pm 40.79$ & university campus & drone camera \\
		sdd:quad00 & $9$ & $27.00 \pm 14.99$ & university campus & drone camera \\
		sdd:quad01 & $14$ & $36.86 \pm 11.12$ & university campus & drone camera \\
		sdd:quad02 & $13$ & $39.54 \pm 7.47$ & university campus & drone camera \\
		sdd:quad03 & $6$ & $26.17 \pm 14.44$ & university campus & drone camera \\
		\hline
	\end{tabular}
	\caption{Details of pedestrian trajectory datasets used in this paper. As the \emph{BIWI}, \emph{crowds} and \emph{Stanford Drone} datasets consist of several scenes and potentially multiple recordings, the (sub-)datasets are denoted as \emph{'DatasetName':'SceneName''RecordingNumber'}.} 
	\label{tab:dataset_details}
\end{table*}

\bibliographystyle{IEEEtran}
\bibliography{bibliography}

\begin{thebibliography}{10}
\providecommand{\url}[1]{#1}
\csname url@samestyle\endcsname
\providecommand{\newblock}{\relax}
\providecommand{\bibinfo}[2]{#2}
\providecommand{\BIBentrySTDinterwordspacing}{\spaceskip=0pt\relax}
\providecommand{\BIBentryALTinterwordstretchfactor}{4}
\providecommand{\BIBentryALTinterwordspacing}{\spaceskip=\fontdimen2\font plus
\BIBentryALTinterwordstretchfactor\fontdimen3\font minus
  \fontdimen4\font\relax}
\providecommand{\BIBforeignlanguage}[2]{{%
\expandafter\ifx\csname l@#1\endcsname\relax
\typeout{** WARNING: IEEEtran.bst: No hyphenation pattern has been}%
\typeout{** loaded for the language `#1'. Using the pattern for}%
\typeout{** the default language instead.}%
\else
\language=\csname l@#1\endcsname
\fi
#2}}
\providecommand{\BIBdecl}{\relax}
\BIBdecl

\bibitem{kalman1960}
R.~Kalman, ``A new approach to linear filtering and prediction problems,''
  \emph{Journal of basic Engineering}, vol.~82, no.~1, pp. 35--45, 1960.

\bibitem{helbing1995social}
D.~Helbing and P.~Molnar, ``Social force model for pedestrian dynamics,''
  \emph{Physical review E}, vol.~51, no.~5, p. 4282, 1995.

\bibitem{becker2018red}
S.~Becker, R.~Hug, W.~H{\"u}bner, and M.~Arens, ``Red: A simple but effective
  baseline predictor for the trajnet benchmark,'' in \emph{The European
  Conference on Computer Vision (ECCV) Workshops}, September 2018.

\bibitem{Nikhil_ECCVW_2018}
N.~Nikhil and B.~Morris, ``Convolutional neural network for trajectory
  prediction,'' in \emph{The European Conference on Computer Vision (ECCV)
  Workshops}, 2018.

\bibitem{gupta2018social}
A.~Gupta, J.~Johnson, L.~Fei-Fei, S.~Savarese, and A.~Alahi, ``Social gan:
  Socially acceptable trajectories with generative adversarial networks,'' in
  \emph{Proceedings of the IEEE Conference on Computer Vision and Pattern
  Recognition}, 2018, pp. 2255--2264.

\bibitem{giuliari2020transformer}
F.~Giuliari, I.~Hasan, M.~Cristani, and F.~Galasso, ``Transformer networks for
  trajectory forecasting,'' in \emph{International Conference on Pattern
  Recognition (ICPR)}, 2020.

\bibitem{alahi2016social}
A.~Alahi, K.~Goel, V.~Ramanathan, A.~Robicquet, L.~Fei-Fei, and S.~Savarese,
  ``Social lstm: Human trajectory prediction in crowded spaces,'' in
  \emph{Proceedings of the IEEE conference on computer vision and pattern
  recognition}, 2016, pp. 961--971.

\bibitem{huang2019stgat}
Y.~Huang, H.~Bi, Z.~Li, T.~Mao, and Z.~Wang, ``Stgat: Modeling spatial-temporal
  interactions for human trajectory prediction,'' in \emph{Proceedings of the
  IEEE/CVF International Conference on Computer Vision}, 2019, pp. 6272--6281.

\bibitem{sadeghian2019sophie}
A.~Sadeghian, V.~Kosaraju, A.~Sadeghian, N.~Hirose, H.~Rezatofighi, and
  S.~Savarese, ``Sophie: An attentive gan for predicting paths compliant to
  social and physical constraints,'' in \emph{Proceedings of the IEEE/CVF
  Conference on Computer Vision and Pattern Recognition}, 2019, pp. 1349--1358.

\bibitem{bisagno2021embedding}
N.~Bisagno, C.~Saltori, B.~Zhang, F.~G. De~Natale, and N.~Conci, ``Embedding
  group and obstacle information in lstm networks for human trajectory
  prediction in crowded scenes,'' \emph{Computer Vision and Image
  Understanding}, vol. 203, p. 103126, 2021.

\bibitem{rudenko2019human}
A.~Rudenko, L.~Palmieri, M.~Herman, K.~Kitani, D.~Gavrila, and K.~Arras,
  ``Human motion trajectory prediction: {A} survey,'' \emph{The International
  Journal of Robotics Research}, vol.~39, no.~8, pp. 895--935, 2020.

\bibitem{rasouli2020deep}
A.~Rasouli, ``Deep learning for vision-based prediction: A survey,'' 2020.

\bibitem{kothari2020human}
P.~Kothari, S.~Kreiss, and A.~Alahi, ``Human trajectory forecasting in crowds:
  A deep learning perspective,'' \emph{arXiv preprint arXiv:2007.03639}, 2020.

\bibitem{sadeghiankosaraju2018trajnet}
A.~Sadeghian, V.~Kosaraju, A.~Gupta, S.~Savarese, and A.~Alahi, ``Trajnet:
  Towards a benchmark for human trajectory prediction,'' \emph{arXiv preprint},
  2018.

\bibitem{robicquet2016learning}
A.~Robicquet, A.~Sadeghian, A.~Alahi, and S.~Savarese, ``Learning social
  etiquette: Human trajectory understanding in crowded scenes,'' in
  \emph{European conference on computer vision}.\hskip 1em plus 0.5em minus
  0.4em\relax Springer, 2016, pp. 549--565.

\bibitem{zhang2020srlstm}
P.~{Zhang}, J.~{Xue}, P.~{Zhang}, N.~{Zheng}, and W.~{Ouyang}, ``Social-aware
  pedestrian trajectory prediction via states refinement lstm,'' \emph{IEEE
  Transactions on Pattern Analysis and Machine Intelligence}, pp. 1--1, 2020.

\bibitem{hug2017reliability}
R.~Hug, S.~Becker, W.~H{\"u}bner, and M.~Arens, ``On the reliability of
  lstm-mdl models for pedestrian trajectory prediction,'' \emph{VIIth
  International Workshop on Representation, analysis and recognition of shape
  and motion FroM Image data (RFMI 2017)}, 2017.

\bibitem{maaten2008tsne}
L.~v.~d. Maaten and G.~Hinton, ``Visualizing data using t-sne,'' \emph{Journal
  of machine learning research}, vol.~9, no. Nov, pp. 2579--2605, 2008.

\bibitem{jolliffe2016pca}
I.~T. Jolliffe and J.~Cadima, ``Principal component analysis: a review and
  recent developments,'' \emph{Philosophical Transactions of the Royal Society
  A: Mathematical, Physical and Engineering Sciences}, vol. 374, no. 2065, p.
  20150202, 2016.

\bibitem{hug2020shortnote}
R.~Hug, S.~Becker, W.~H{\"u}bner, and M.~Arens, ``A short note on analyzing
  sequence complexity in trajectory prediction benchmarks,'' \emph{2nd Workshop
  on Long-term Human Motion Prediction}, 2020.

\bibitem{amirian2020opentraj}
J.~Amirian, B.~Zhang, F.~V. Castro, J.~J. Baldelomar, J.-B. Hayet, and
  J.~Pettre, ``Opentraj: Assessing prediction complexity in human trajectories
  datasets,'' in \emph{Proceedings of the Asian Conference on Computer Vision},
  2020.

\bibitem{lloyd1982least}
S.~Lloyd, ``Least squares quantization in pcm,'' \emph{IEEE transactions on
  information theory}, vol.~28, no.~2, pp. 129--137, 1982.

\bibitem{ester1996density}
M.~Ester, H.-P. Kriegel, J.~Sander, X.~Xu \emph{et~al.}, ``A density-based
  algorithm for discovering clusters in large spatial databases with noise.''
  in \emph{Kdd}, vol.~96, no.~34, 1996, pp. 226--231.

\bibitem{kohonen2001lvq}
T.~Kohonen, M.~R. Schroeder, and T.~S. Huang, \emph{Self-Organizing Maps},
  3rd~ed.\hskip 1em plus 0.5em minus 0.4em\relax Berlin, Heidelberg:
  Springer-Verlag, 2001.

\bibitem{xie2016unsupervised}
J.~Xie, R.~Girshick, and A.~Farhadi, ``Unsupervised deep embedding for
  clustering analysis,'' in \emph{International conference on machine
  learning}.\hskip 1em plus 0.5em minus 0.4em\relax PMLR, 2016, pp. 478--487.

\bibitem{caron2018deep}
M.~Caron, P.~Bojanowski, A.~Joulin, and M.~Douze, ``Deep clustering for
  unsupervised learning of visual features,'' in \emph{Proceedings of the
  European Conference on Computer Vision (ECCV)}, 2018, pp. 132--149.

\bibitem{peng2019deep}
X.~Peng, H.~Zhu, J.~Feng, C.~Shen, H.~Zhang, and J.~T. Zhou, ``Deep clustering
  with sample-assignment invariance prior,'' \emph{IEEE transactions on neural
  networks and learning systems}, vol.~31, no.~11, pp. 4857--4868, 2019.

\bibitem{xu2015comprehensive}
D.~Xu and Y.~Tian, ``A comprehensive survey of clustering algorithms,''
  \emph{Annals of Data Science}, vol.~2, no.~2, pp. 165--193, 2015.

\bibitem{min2018survey}
E.~Min, X.~Guo, Q.~Liu, G.~Zhang, J.~Cui, and J.~Long, ``A survey of clustering
  with deep learning: From the perspective of network architecture,''
  \emph{IEEE Access}, vol.~6, pp. 39\,501--39\,514, 2018.

\bibitem{oord2017neural}
A.~van~den Oord, O.~Vinyals \emph{et~al.}, ``Neural discrete representation
  learning,'' in \emph{Advances in Neural Information Processing Systems},
  2017, pp. 6306--6315.

\bibitem{quehl2017good}
J.~Quehl, H.~Hu, {\"O}.~{\c{S}}. Ta{\c{s}}, E.~Rehder, and M.~Lauer, ``How good
  is my prediction? finding a similarity measure for trajectory prediction
  evaluation,'' in \emph{2017 IEEE 20th International Conference on Intelligent
  Transportation Systems (ITSC)}.\hskip 1em plus 0.5em minus 0.4em\relax IEEE,
  2017, pp. 1--6.

\bibitem{pena1999empirical}
J.~M. Pena, J.~A. Lozano, and P.~Larranaga, ``An empirical comparison of four
  initialization methods for the k-means algorithm,'' \emph{Pattern recognition
  letters}, vol.~20, no.~10, pp. 1027--1040, 1999.

\bibitem{pellegrini2009biwi}
S.~Pellegrini, A.~Ess, K.~Schindler, and L.~van Gool, ``You'll never walk
  alone: Modeling social behavior for multi-target tracking,'' in
  \emph{International Conference on Computer Vision}, 2009, pp. 261--268.

\bibitem{lerner2007crowds}
A.~Lerner, Y.~Chrysanthou, and D.~Lischinski, ``Crowds by example,''
  \emph{Computer Graphic Forum}, vol.~26, no.~3, pp. 655--664, 2007.

\bibitem{hasan2019forecasting}
I.~{Hasan}, F.~{Setti}, T.~{Tsesmelis}, V.~{Belagiannis}, S.~{Amin}, A.~{Del
  Bue}, M.~{Cristani}, and F.~{Galasso}, ``Forecasting people trajectories and
  head poses by jointly reasoning on tracklets and vislets,'' \emph{IEEE
  Transactions on Pattern Analysis and Machine Intelligence}, pp. 1--1, 2019.

\end{thebibliography}

\begin{IEEEbiography}[{\includegraphics[width=1in,height=1.25in,clip,keepaspectratio]{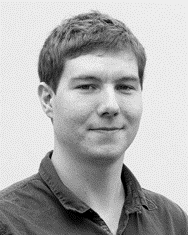}}]{Ronny Hug} 
	received his master's degree in computer science from the Karlsruhe Institute of Technology (KIT) in 2015.
	Since 2015, he works as a research assistant at the Fraunhofer Institute for Optronics, System Technologies and Image Exploitation (IOSB) in the \emph{Video Content Analysis} group.
	His research mainly revolves around probabilistic sequence modeling, with a focus on human trajectory modeling.
\end{IEEEbiography}

\begin{IEEEbiography}[{\includegraphics[width=1in,height=1.25in,clip,keepaspectratio]{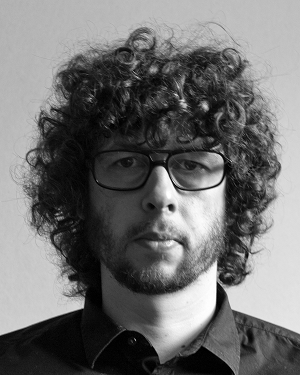}}]{Stefan Becker} 
	received his Ph.D. (Dr.-Ing) in computer science and his diploma in electric engineering from the Karlsruhe Institute of Technology (KIT). 
	In 2011, he joined the Fraunhofer Institute for Optronics, System Technologies, and Image Exploitation (IOSB) where he is currently working as postdoctoral researcher in the \emph{Video Content Analysis} group. 
	He participated and contributed to several projects in industry, government, and EU.
\end{IEEEbiography}

\begin{IEEEbiography}[{\includegraphics[width=1in,height=1.25in,clip,keepaspectratio]{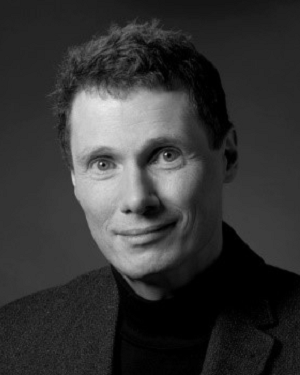}}]{Wolfgang H\"ubner} 
	is head of the \emph{Video Content Analysis} group at the Fraunhofer Institute for Optronics, System Technologies, and Image Exploitation (IOSB). 
	He received his Ph.D. in computer science in 2005 from the University of Bremen. 
	Before and after that he worked at the Cognitive Neuroscience Lab at the University of T\"ubingen on different publicly funded research projects in the fields of e-Learning in cognitive sciences, cognitive robotics, and cognitive vision applications. 
	His main research interests include machine learning applications, especially in the context of cognitive systems, interactive systems and automated signal analysis.
\end{IEEEbiography}

\begin{IEEEbiography}[{\includegraphics[width=1in,height=1.25in,clip,keepaspectratio]{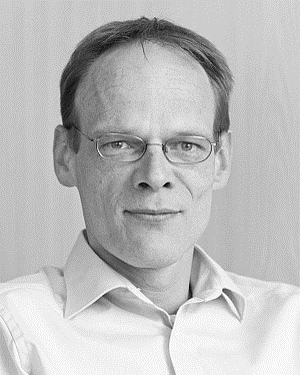}}]{Michael Arens} 
	received his diploma in Computer Science and his PhD (Dr.rer.nat.) from the University of Karlsruhe in 2001 and 2004, respectively. 
	Since 2006, he works at the Fraunhofer Institute of Optronics, System Technologies and Image Exploitation in various positions. 
	In 2011, Dr. Arens was appointed Head of the Department of Object Recognition. 
	His research interests include image and video analysis with an emphasis on the automatic conceptualization and reasoning on results obtained from such an analysis where he has co-authored several articles. 
	Dr. Arens is member of the German Association of Pattern Recognition (DAGM).
\end{IEEEbiography}

\EOD

\end{document}